\documentclass{article}
\usepackage{indentfirst}
\usepackage{amsmath,amscd,amssymb}
\usepackage{algorithm}
\usepackage[noend]{algpseudocode}
\usepackage[margin=1in]{geometry}
\usepackage[all]{nowidow}
\usepackage{hyperref}
\usepackage{natbib}

\usepackage{bbold}
\usepackage{bm}
\usepackage{rotating,graphicx}
\usepackage{amsfonts,amsmath,amscd,amssymb}
\usepackage{cancel}

\newcommand{\Ccal}{\mathcal{C}}
\newcommand{\Rcal}{\mathcal{R}}
\newcommand{\Dcal}{\mathcal{D}}
\newcommand{\Lcal}{\mathcal{L}}
\newcommand{\yvec}{\mathbf{y}}
\newcommand{\xvec}{\mathbf{x}}
\newcommand{\wvec}{\mathbf{w}}
\newcommand{\dd}{\mathrm{d}}
\newcommand{\muvec}{\pmb \mu}
\newcommand{\epsvec}{\pmb \epsilon}
\newcommand{\sigmavec}{\pmb \sigma}
\newcommand{\sigmasvec}{\pmb \sigma^2}
\newcommand{\norm}{\mathcal{N}}
\newcommand{\diag}{\mathrm{diag}}
\newcommand{\zvec}{\mathbf{z}}
\newcommand{\Vvec}{\mathbf{V}}
\newcommand{\uvec}{\mathbf{u}}
\newcommand{\cvec}{\mathbf{c}}

\newcommand{\Avec}{\mathbf{A}}
\newcommand{\Bvec}{\mathbf{B}}
\newcommand{\Kfu}{\mathbf{K}_{\wvec\uvec}}
\newcommand{\Kuf}{\mathbf{K}_{\uvec\wvec}}
\newcommand{\Kuu}{\mathbf{K}_{\uvec\uvec}}
\newcommand{\Kff}{\mathbf{K}_{\wvec\wvec}}

\newcommand{\Cvec}{\mathbf{C}}

\usepackage{hyperref}
\usepackage[capitalise]{cleveref}

\usepackage{authblk}

\begin{document}

\author{
  Theofanis $\text{Karaletsos}^{*}$, Thang D. $\text{Bui}^{*}$\\
  \texttt{theofanis@uber.com, thang.bui@uber.com}
}
\affil{Uber AI, San Francisco, California, USA}
\affil[*]{authors contributed equally}

\title{Hierarchical Gaussian Process Priors for Bayesian Neural Network Weights}

\date{}

\maketitle

\begin{abstract}
Probabilistic neural networks are typically modeled with independent weight priors, which do not capture weight correlations in the prior and do not provide a parsimonious interface to express properties in function space. 
A desirable class of priors would represent weights compactly, capture correlations between weights, facilitate calibrated reasoning about uncertainty, and allow inclusion of prior knowledge about the function space such as periodicity or dependence on contexts such as inputs.
To this end, this paper introduces two innovations: (i) a Gaussian process-based hierarchical model for network weights based on unit embeddings that can flexibly encode correlated weight structures, and (ii) input-dependent versions of these weight priors that can provide convenient ways to regularize the function space through the use of kernels defined on contextual inputs. We show these models provide desirable test-time uncertainty estimates on out-of-distribution data, demonstrate cases of modeling inductive biases for neural networks with kernels which help both interpolation and extrapolation from training data, and demonstrate competitive predictive performance on an active learning benchmark.

\end{abstract}

\section{Introduction}
\label{sec:background}

Bayesian neural networks (BNNs) \citep[see e.g.][]{mackay:92,neal1992bayesian,ghahramani2016history} are one of the research frontiers on combining Bayesian inference and deep learning, potentially offering flexible modelling power with calibrated predictive performance. In essence, applying probabilistic inference to neural networks allows all plausible network parameters, not just the most likely, to be used for predictions. Despite the strong interest in the community for the exploration of BNNs, there remain unanswered questions: (i) how can we model neural network functions to encourage behaviors such as interpolation between signals and extrapolation from data in meaningful ways, for instance by encoding prior knowledge, or how to specify priors which facilitate uncertainty quantification, and (ii) many scalable approximate inference methods are not rich enough to capture complicated posterior correlations in large networks, resulting in undesirable predictive performance at test time.

This paper attempts to tackle some of the aforementioned limitations. We propose inherently correlated weight priors by utilizing unit-level latent variables to obtain a compact parameterization of neural network weights and combine them with ideas from the Gaussian process (GP) literature to induce a hierarchical GP prior over weights. This prior flexibly models the correlations between weights in a layer and across layers. We explore the use of product kernels to implement input-dependence as a variation of the proposed prior, yielding models that have per-datapoint priors which facilitate inclusion of prior knowledge through kernels while maintaining their weight structure. A structured variational inference approach is employed that side-steps the need to do inference in the weight space whilst retaining weight correlations in the approximate posterior. The proposed priors and approximate inference scheme are demonstrated to exhibit beneficial properties for tasks such as generalization, uncertainty quantification, and active learning.

The paper is organized as follows: in \cref{sec:metarep_nn} we review hierarchical modeling for BNNs based on unit-variables. In \cref{sec:metarep_gp} we introduce the global and local weight models and their applications to neural networks.
Efficient inference algorithms for both models are presented in \cref{sec:inference}, followed by a suite of experiment to validate their performance in \cref{sec:exp}. We review related work in \cref{sec:related}. 

\section{Meta-representing weights and networks}
\label{sec:metarep_nn}
Our work builds on and expands the class of hierarchical neural network models based on the concept of latent variables associated with units in a network as proposed in \citep{karaletsos2018probabilistic}.
In that model, each unit (visible or hidden) of the $l$-th layer of the network has a corresponding latent hierarchical variable $\zvec_{l, i}$,  of dimensions $D_z$, where $i$ denotes the index of the unit in a layer. Note that these latent variables do not describe the activation of units, but rather constitute latent features associated with a unit.

The design of these latent variables is judiciously chosen to construct the weights in the network as follows: a weight in the $l$-th layer, $w_{l,i,j}$ is generated by using the concatenation of latent variable $z$'s of the $i$-th input unit and the $j$-th output unit as inputs of a mapping function $f(\big[ {\bf z}_{l,i}, {\bf z}_{l+1,j}\big])$.

We can summarize this relationship by introducing a set of {\it weight encodings} ${\bf C}_{w}(\zvec)$, one for each individual weight in the network ${\bf c}_{w_{l,i,j}} = \big[ {\bf z}_{l,i}, {\bf z}_{l+1,j}\big]$, which can be deterministically constructed from the collection of unit latent variable samples {\bf z} by concatenating them correctly according to network architecture. The probabilistic description of the relationship between the weight codes (summarizing the structured latent variables) and the weights $\wvec$ is:
\begin{align*}
  p(\wvec | \zvec) = p(\wvec | {\bf C}_{w}(\zvec)) = \prod_{l=1}^{L-1} \prod_{i=1}^{H_l} \prod_{j=1}^{H_{l+1}} p(w_{l, i, j}  | {\bf c}_{w_{l,i,j}}),
\end{align*}
where $l$ denotes a visible or hidden layer and $H_l$ is the number of units in that layer, $L$ is the total number of layers in the network, and $\wvec$ denotes all the weights in this network.

In \cite{karaletsos2018probabilistic}, a small parametric neural network regression model (conceptually a {\it structured hyper-network}) is chosen to map the latent variables to the weights, using either a Gaussian noise model $p(\wvec|{\bf C}_{w}(\zvec), \theta) = \mathcal{N}(\wvec| \mathrm{NN}_\theta({\bf C}_{w}(\zvec))$
or an implicit noise model: 
$p(\wvec|{\bf C}_{w}(\zvec), \theta) \propto \mathrm{NN}_\theta({\bf C}_{w}(\zvec),\epsilon )$,
where $\epsilon$ is a random variate.
We will call this network a {\it meta mapping}.
Note that given the collection of sampled unit variables $\zvec$ and the derived codes ${\bf C}_{w}(\zvec)$, the weights (or theirs mean and variance) can be obtained efficiently in parallel. 
A prior over latent variables $\zvec$ completes the model specification, $p(\zvec) = \prod_{l=0}^{L}\prod_{i = 0}^{H_l} \norm(\zvec_{l,i}; \mathbf{0}, \mathbf{I})$.
The joint density of the resulting hierarchical BNN is then specified as follows,
\begin{align*}
p({\bf y}, \wvec, \zvec |{\bf x}, \theta)
  = p(\zvec) p(\wvec | {\bf C}_{w}(\zvec), \theta) \prod_{n=1}^N p(\yvec_n | \wvec, \xvec_n),
\end{align*}
with $N$ denoting the number of observation tuples $\{\xvec_n, \yvec_n\}$.

Variational inference was employed in prior work to infer $\zvec$ (and $\wvec$ implicitly), and to obtain a point estimate of $\theta$, as a by-product of optimising the variational lower bound. Critically, in this representation weights are only implicitly parametrized through the use of these latent variables, which transforms inference on weights into inference of the much smaller collection of latent unit variables.

The central motivations for our adoption of this parameterization are two-fold. {\it First}, the number of visible and hidden units in a neural network is typically much smaller than the number of weights. For example, for the $l$-th weight layer, there are $H_{l} \times H_{l+1}$ weights compared to $H_{l} + H_{l+1}$ associated latent variables. This encourages the development of models and inference schemes that can work in the lower-dimensional hierarchical latent space directly without the need to model individual weights privately.
This structured representation per weight allows powerful hierarchical models to be used without requiring high dimensional parametrizations (i.e. hypernetworks for the entire weight tensor). Specifically, a GP-LVM prior~\cite{lawrence2004gaussian} over all network weights appears infeasible without such an encoding of structure.
{\it Second}, the compact latent space representation facilitates attempts at building fine-grained control into weight priors, such as structured prior knowledge as we will see in the following sections.
\begin{figure}[!ht]
    \centering
    \includegraphics[width=0.8\columnwidth]{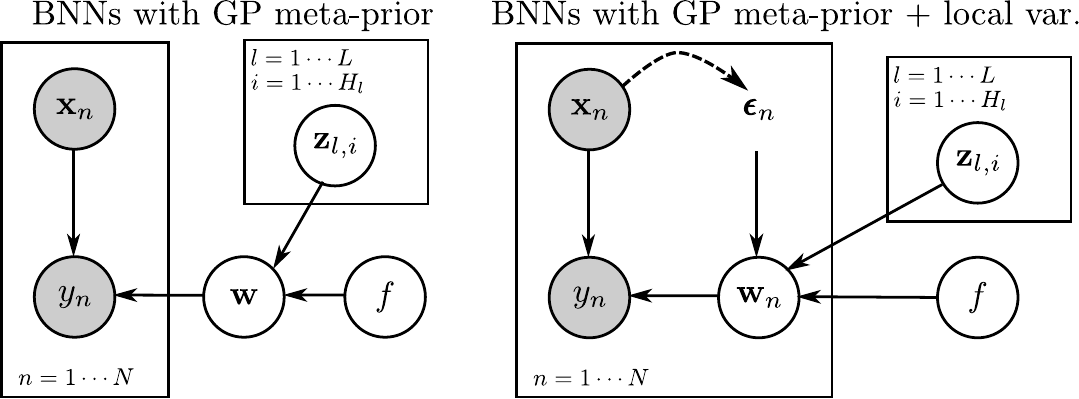}
    \caption{Graphical depiction of  BNNs with hierarchical GP-MetaPriors and ones with input-dependent variables.}
    \label{fig:models}
\end{figure}

\section{Hierarchical GP-Priors For BNN Weights}
\label{sec:metarep_gp}
Notice that in \Cref{sec:metarep_nn}, the meta mapping from the hierarchical latent variables to the weights is a parametric non-linear function, specified by a neural network. We replace the parametric neural network by a probabilistic functional mapping and place a nonparametric Gaussian process prior over this function. That is,
\begin{align*}
  p(w_{l, i, j}  | f, {\bf c}_{w_{l,i,j}}) &= \norm(w_{l, i, j}; f([\zvec_{l, i}, \zvec_{l+1, j}]), \sigma_w^2), \\
  p(f|\gamma) &= \mathcal{GP}(f; \mathbf{0}, k_{w}(\cdot, \cdot|\gamma)),
\end{align*}
where we have assumed a zero-mean GP, $k_\gamma(\cdot,\cdot)$ is a covariance function and $\gamma$ is a small set of hyper-parameters, and a homoscedastic\footnote{Here, we present a homoscedastic noise model for the weights, but the model is readily adaptable to a heteroscedastic noise model which we omit for clarity.} Gaussian noise model with variance $\sigma_{w}^{2}$.

The effect is that the latent function introduces correlations for the individual weight predictions,
\begin{align*}
\begin{split}
P(\wvec|{\bf z}) &= P(\wvec|{\bf C}_w({\bf z}))\\ &= \int_f p(f) \Big[ \prod \limits_{l=1}^{L-1} \prod \limits_{i=1}^{H_{1}} \prod \limits_{j=1}^{H_{l+1}} p(w_{l,i,j}|f, \zvec_{l,i}, \zvec_{l+1,j})\Big].
\end{split}
\end{align*}

Notably, while the number of latent variables and weights can be large, the input dimension to the GP mapping is only $2D_z$, where $D_z$ is the dimensionality of each latent variable $\zvec$. 
The GP mapping effectively performs one-dimensional regression from latent variables to individual weights while capturing their correlations. 
We will refer to this mapping as a {\bf GP-MetaPrior} ({\it MetaGP}). We define the following kernel at the example of two weights in the network,
\begin{equation*}
\begin{split}
k_{w}(c_{w_{1}}, c_{w_{2}}) &= k_{w}([\zvec_{l^{1}, i^{1}}, \zvec_{l^{1}+1, j^{1}}],[\zvec_{l^{2}, i^{2}}, \zvec_{l^{2}+1, j^{2}}])
\end{split}
\end{equation*}

In this section and what follows, we will use the popular exponentiated quadratic (RBF) kernel with ARD lengthscales, $k(\xvec_{1}, \xvec_2) = \sigma_k^2 \exp\left(\sum_{d=1}^{2D_z} \frac{-(x_{1,d} - x_{2, d})^2}{2l_d^2}\right)$, where $\{l_{d}\}_{d=1}^{2D_z}$ are the lengthscales and $\sigma_k^2$ is the kernel variance. 

{\bf BNNs with GP-MetaPriors} are then specified by the following joint density over all variables:
\begin{align}
  p({\bf y}, \wvec, \zvec, f | {\bf x}) 
  &= p(\zvec) p(f) p(\wvec | f, \zvec) p(\yvec | \wvec, \xvec) \nonumber\\
  &= p(\zvec) p(f) p(\wvec|f,{\bf C}_{w}({\bf z}))\prod_{n=1}^N \left[ p(\yvec_n | \wvec, \xvec_n) \right]. \nonumber
\end{align}

\begin{figure}[!ht]
    \centering
    \includegraphics[width=0.9\columnwidth]{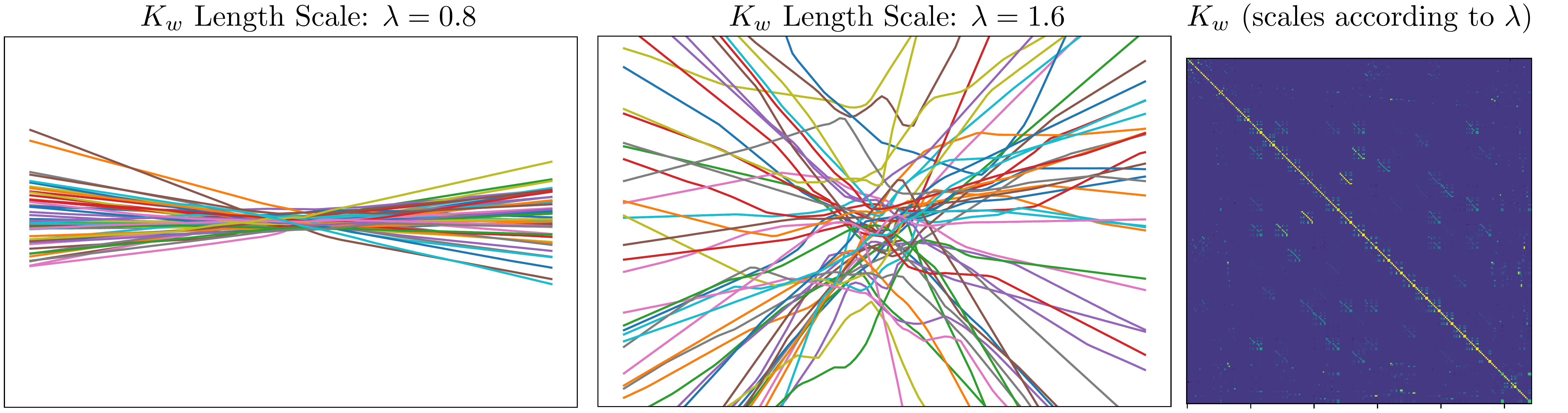}
    \caption{{\bf MetaGP Prior Samples}: We show function samples generated from a [1,20,10,1] unit BNN with ReLUs with meta-GP prior. We draw one sample per unit ${\bf z}_{\mathrm{viz}} \sim p({\bf z})$ to instantiate the weight prior and subsequently draw 40 function samples (individually colored) from the BNN by drawing from the conditional prior ${\bf w} \sim \mathcal{N}({\bf w}|{\bf z}_{\mathrm{viz}})$ and regressing ${\bf y} \sim p({\bf y} |{\bf w} ,{\bf x})$. ({\bf left}) We show samples drawn when the global RBF kernel for the GP has a length scale set to a small value. ({\bf middle}) we keep the same samples {\bf z} and only change the length scale to be larger and visualize the functions induced by the BNN ({\bf right}) We visualize the weight kernel given the latent variables ${\bf z}_{\mathrm{viz}}$. Other samples of {\bf z} would induce different weight covariance matrices. Overall this figure shows that even given {\bf z}, the proposed prior models a wide range of functions which have controllable properties based on the parameters of the kernel.}
    \label{fig:metagp-samples}
\end{figure}

We show prior samples from this model in \cref{fig:metagp-samples} by the following procedure: for a sample ${\bf z}_{\mathrm{viz}}\sim p(\zvec)$ we instantiate the covariance matrix ${\bf K}_w$ by constructing weight codes and applying the kernel function $k_w$. We draw weights from the GP by sampling the Normal distribution $\mathcal{N}(\wvec;\mathbf{0},{\bf K}_w + \sigma_{w}^{2} \mathbf{I})$, where ${\bf K}_w \in \mathcal{R}^{|\wvec| \times |\wvec|}$, with $|\wvec|$ denoting the number of all parameters in the network. We then generate BNN function samples given the sampled weights with homoscedastic noise on the outputs. We highlight two properties of this model: {\it First}, as a hierarchical model, given a sample of the latent variables {\bf z}, the model instantiates a prior over weights from which we can further sample functions and thus encodes two levels of uncertainty over functions. {\it Second}, we demonstrate that changing the length-scale parameter of the mapping-kernel $k_{w}$, again even {\it given fixed samples {\bf z}}, leads to vastly differentiated function samples, showing the compact degree of control the mapping parameters have over the function space being modeled. In this case, the length-scale appears to control the variance over the function space, which matches an intuitive interpretation over the kernel parameter.

An important task is marginalization of the latent quantities given data to perform posterior inference, which we discuss in \cref{sec:inference_global}.

\subsection{Input-kernels for modulating function priors}
\label{sec:local_model}
While the hierarchical latent variables and meta mappings introduce non-trivial coupling between the weights a priori, they are inherently global. That is, a function drawn from the model, represented by a set of weights, does not take into account the inputs at which the function will be evaluated. In this section, we will describe modifications to our weight prior which allow conditional weight models on inputs.

To this end, we introduce the input variable into the weight codes ${\bf c}_{w_{n,l,i,j}}=
\big[ {\bf c}_{w_{l,i,j}}, {\bf x}_n \big]=\big[\zvec_{l,i}, \zvec_{l+1,j}, {\bf x}_n \big]$, which we utilize to yield input-conditional weight models $p(w_{n, l, i, j}  | f, \zvec_{l, i}, \zvec_{l+1, j}, \xvec_n)$ through the use of product kernels. Concretely, we introduce a new {\bf input kernel} $k_{\mathrm{aux}}$ which multiplied with the global weight kernel $k_w$ gives the kernel $k_{\mathrm{local}}$ for the meta mapping,
\begin{align*}
\begin{split}
k_{\mathrm{local}}({\bf c}_{w_{1},x_{1}}, {\bf c}_{w_{2},x_{2}}) &=
k_{w}({\bf c}_{w_{1}}, {\bf c}_{w_{2}}) \cdot k_{\mathrm{aux}}({\bf x}_{1}, {\bf x}_{2})
\end{split}
\end{align*}
where $k_w$ is the kernel defined over latent-variable weight codes from \cref{sec:metarep_gp}, $k_{\mathrm{aux}}$ is an auxiliary kernel modeling input-dependence on ${\bf x}_n$, and ${\bf c}_{w_{l,i,j},x_{n}}
$ is shorthand for ${\bf c}_{w_{n,l,i,j}}$.
This factorization over kernels represents an assumption of separable influence on functions by latent variables $\zvec$ and inputs. The weight priors are now also local to each data point, in a similar vein to how functions are drawn from a GP, while still instantiating an explicit, weight-based model.

\begin{figure}[!ht]
    \centering
    \includegraphics[width=.8\columnwidth]{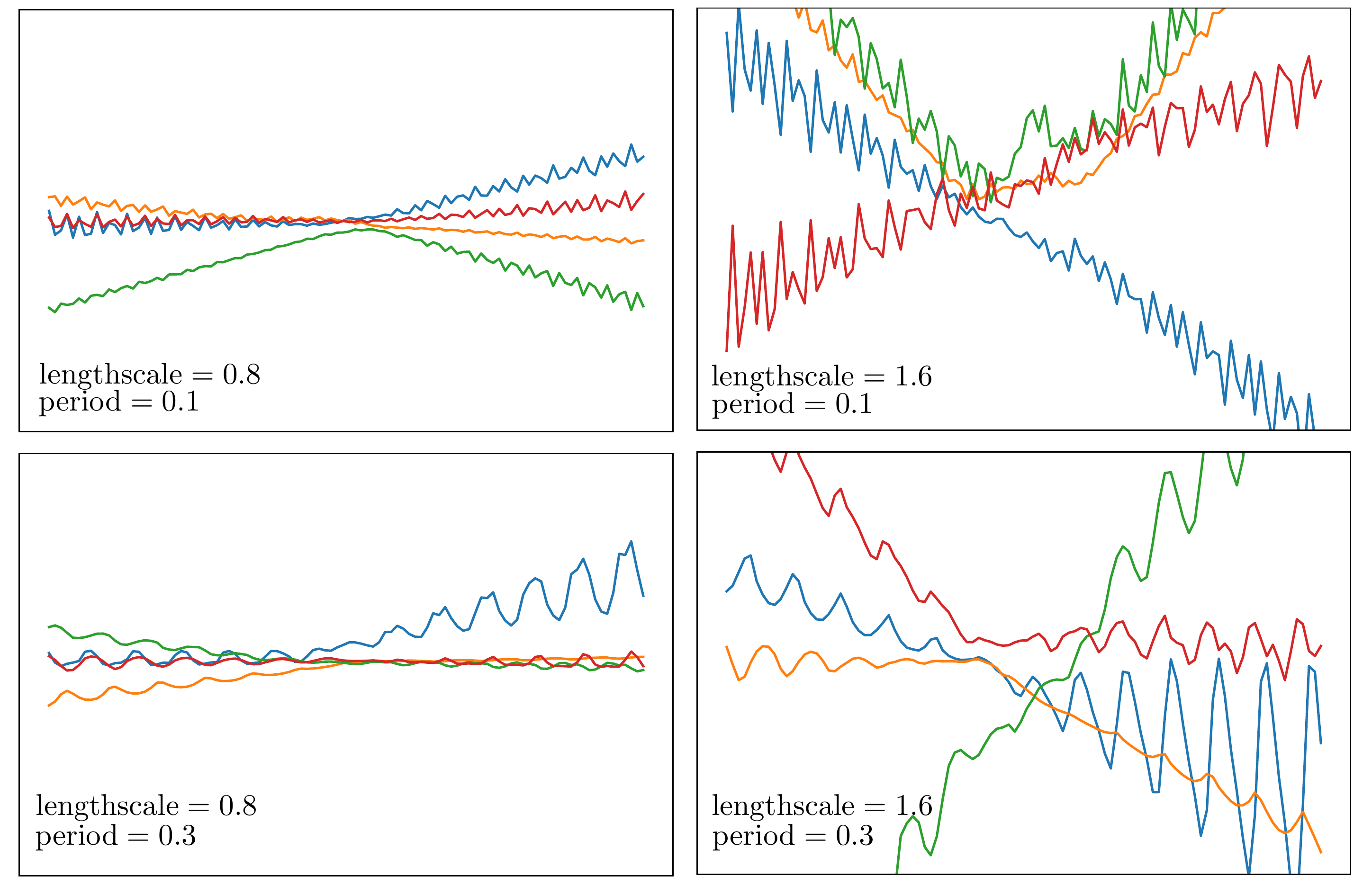}
    \caption{{\bf Local MetaGP Samples}: 
    We remind the reader that the input-dependent weight prior has a factorized kernel structure $k_{\mathrm{local}} = k_{w} \cdot k_{\mathrm{aux}}$, and we wish to demonstrate the effect of each kernel separately in terms of its effects on the induced function prior for the neural network. We are given the same samples {$\bf z_{\mathrm{viz}}$} as in \cref{fig:metagp-samples} and also keep the two kernel parameter choices for $k_w$, while varying only the period parameter for an auxiliary periodic kernel. {\bf Left}: We show function samples using a small period of 0.1 and a period of 0.3 in combination with the $k_w$ kernel with length-scale 0.8. We can see, that while the functions are still relatively flat, the auxiliary kernel induces weight priors which lead to periodic function samples consistent with the auxiliary kernel setting. {\bf Right}: Similarly, when performing the same protocol for the $k_w$ with the larger lengthscale, we again observe periodic functions consistent with the set period (although we only apply the periodic kernel for the weight priors for the BNN), but see that the functions sampled have more variance, consistent with the larger length-scale of the weight-kernel $k_w$. Note that while the functions exhibit periodic structure, they have non-periodic global structure as well, as they also draw information from $k_w$ and the priors are merely {\it modulated} by the auxiliary kernel. We thus see that our prior structure successfully induces function priors which naturally inherit properties we can express as kernel functions, but keep rich expressivity as weight based models.}
    \label{fig:local_metagp-samples}
\end{figure}

We demonstrate the effects of utilizing the auxiliary kernel in this factorized fashion by visualizing prior function samples from a BNN with this local prior when changing kernel parameters in \cref{fig:local_metagp-samples}, exemplifying the proposed model's ability to encode controlled periodic structure into BNN weight priors before seeing any data. As performed in~\cref{sec:metarep_gp}, we sample from the GP to instantiate weights, but in the case of the local model we instantiate the covariance matrix ${\bf K}_{\mathrm{local}} \in \mathcal{R}^{|\wvec| \times |\wvec| \times N \times N}$. We will discuss the handling of this conceptually large object in~\cref{sec:inference_local}.

To scale this to large inputs, we learn transformations of inputs for the conditional weight model $\epsvec_n = g(\Vvec \xvec_n)$, for a learned mapping $\Vvec$ and a nonlinearity $g$ and generalize weight codes to ${\bf c}_{w_{n,l,i,j}}=\big[\zvec_{l,i}, \zvec_{l+1,j}, {\bf \epsilon}_n \big]$, with ${\bf C}_{w,x}(\zvec,\xvec)$ describing their collection.
In detail, each auxiliary input is obtained via a (potentially nonlinear) transformation applied to an input: $\epsvec_n = g(\Vvec \xvec_n)$, where $\Vvec \in \Rcal^{D_{\mathrm{aux}} \times D_x}$, and $D_{\mathrm{aux}}$ and $D_x$ are the dimensionality of $\epsvec_n$ and $\xvec_n$, respectively, and $g(\cdot)$ is an arbitrary transformation. We may also layer these transformations in general. We typically set $D_{\mathrm{aux}} \ll D_x$ so this transformation could be thought of as a dimensionality reduction operation. For low dimensional inputs, we set $\epsvec_n = \xvec_n$.

Including these transformations yields the weight model $p(w_{n, l, i, j}  | f, \zvec, \Vvec, \xvec_n) = \norm(w_{n, l, i, j}; f({\bf c}_{w_{n,l,i,j}}), \sigma_w^2)$,
that is, the input dimension of the meta mapping is now $2D_z + D_{\mathrm{aux}}$. Additionally, we also place a prior over the linear transformation: $p(\Vvec) = \norm(\Vvec; \mathbf{0}, \mathbf{I})$. 
We will refer to this mapping as a {\bf Local GP-MetaPrior} ({\it MetaGP-local}).

{\it What effects should we expect from such a modulation?} Consider the use of an exponentiated quadratic kernel: we would expect data which lies far away from training data to receive small kernel values from $K_{aux}$. This, in turn, would modulate the entire kernel $K_{local}$ for that data point to small values, leading to a weight model that reverts increasingly to the prior. We would expect such a model to help with modeling uncertainty by resetting weights to uninformative distributions away from training data.
One may also want to use this mechanism to express inductive biases about the function space, such as adding structure to the weight prior that can be captured with a kernel. This is an appealing avenue, as multiple useful kernels have been found in the GP literature that allow modelers to describe relationships between data, but have previously not been accessible to neural network modelers. We consider this a novel form of functional regularization through the weight prior, which can imbue the entire network with structure that will constrain its function space. 

{\bf BNNs with Local GP-MetaPriors} specify neural networks with individual weight priors per datapoint (also see Graphical Model in \cref{fig:models}):
\begin{align}
  p(\cdot) 
  &= p(\zvec) p(f) \prod_{n=1}^{N} p(\yvec_{n}, \wvec_{n} | f, \zvec, \xvec_{n}) \nonumber\\
  &= p(\zvec) p(f) \prod_{n=1}^{N} p(\wvec_{n}|f,{\bf C}_{w}({\bf z}, \xvec_{n}))\left[ p(\yvec_n | \wvec_{n}, \xvec_n) \right]. \nonumber
\end{align}

Inference and learning are modified accordingly as explained in \cref{sec:inference_local}.

\section{Inference and learning using stochastic structured variational inference}
\label{sec:inference}
Performing inference is challenging due to the non-linearity of the neural network and the need to infer an entire latent function $f$. 
In~\cref{sec:inference_global} we address these problems for {\it MetaGP}, deriving a structured variational inference scheme that makes use of innovations from inducing point GP approximation literature \citep{titsias2009variational,hensman2013gaussian,quinonero2005unifying,matthews2016sparse,bui2017unifying} and previous work on inferring meta-representations \citep{karaletsos2018probabilistic}.
In~\cref{sec:inference_local} we will highlight the modifications necessary to make this inference strategy work for {\it MetaGP-local}.

\subsection{Inference for the global model}
\label{sec:inference_global}
A common strategy for variational inference in GPs is the utilization of inducing points, which entails the construction of learned inputs to the function and corresponding function values which jointly take the place of representative data points. The inducing inputs and outputs, $\{\Cvec_\uvec, \uvec\}$, will be used to parameterize the approximation.

We first partition the space  $\Ccal$ of inputs 
(or {\it weight codes}) to the function $f$ into a finite set of $M$ variables called inducing inputs $\Cvec_{\uvec} = \{\cvec_{u, m} \}_{m=1}^{M}$ where $\cvec_{u, m}  \in \mathcal{R}^{2D_z}$  and the remaining inputs, $\Ccal = \{\Cvec_\uvec{}, \Ccal_{\neq \Cvec_\uvec}\}$. The function $f$ is partitioned identically, $f = \{\uvec, f_{\neq \uvec}\}$, where $\uvec = f(\Cvec_\uvec)$. We can then rewrite the GP prior as follows, $p(f) = p(f_{\neq \uvec} | \uvec) p(\uvec)$.\footnote{The conditioning on $\Ccal_{\neq \Cvec_\uvec}$ and $\Cvec_\uvec$ in $p(\uvec)$ and $p(f_{\neq \uvec} | \uvec)$ is made implicit here and in the rest of this paper.} In particular, a variational approximation is judiciously chosen to mirror the form of the joint density:
\begin{align}
  q(\wvec, \zvec, f) = q(\zvec) p(f_{\neq \uvec} | \uvec) q(\uvec) p(\wvec | f, \zvec),
\end{align}
where the variational distribution over $\wvec$ is made to explicitly depend on remaining variables through the conditional prior, and $q(\zvec)$ is chosen to be a diagonal (mean-field) Gaussian density, $q(\zvec) = \norm(\zvec; \muvec_\zvec, \diag(\sigmasvec_\zvec))$, and $q(\uvec)$ is chosen to be a correlated multivariate Gaussian, $q(\uvec) = \norm(\uvec; \muvec_\uvec, \Sigma_\uvec)$. This approximation allows convenient cancellations yielding a tractable variational lower bound as follows,
\begin{align}
  \mathcal{F}(\cdot) 
  &= \int_{q(\wvec, \zvec, f)}  \log \frac{p(\zvec) \cancel{p(f_{\neq \uvec} | \uvec)} p(\uvec) \cancel{p(\wvec | f, \zvec)} p(\yvec | \wvec, \xvec)}{q(\zvec) \cancel{p(f_{\neq \uvec} | \uvec)} q(\uvec) \cancel{p(\wvec | f, \zvec)}} \nonumber\\
  &\approx - \mathrm{KL}[q(\zvec) || p(\zvec)] - \mathrm{KL}[q(\uvec) || p(\uvec)]\nonumber\\ &\qquad\qquad + \frac{1}{S}\sum_{s=1} ^ {S}\int_{\wvec, f} q(\wvec, f | \zvec_s) \log p(\yvec | \wvec, \xvec), \nonumber
\end{align}
where the last expectation has been approximated by simple Monte Carlo with the reparameterization trick, i.e.~$\zvec_s \sim q(\zvec)$ \citep{salimans2013fixed,kingma2013auto,titsias2014doubly}. We will next discuss how to approximate the expectation $\mathcal{F}_s = \int_{\wvec, f} q(\wvec, f | \zvec_s) \log p(\yvec | \wvec, \xvec)$. Note that we split f into $f_{\neq \uvec}$ and $\uvec$, and that we can integrate $f_{\neq \uvec}$ out exactly to give, $q(\wvec | \zvec_s, \uvec) = \norm(\wvec; \Avec^{(s)} \uvec, \Bvec^{(s)})$,
\begin{align*}
\Avec^{(s)} &= \Kfu^{(s)} \Kuu^{-1}, \\
\Bvec^{(s)} &= \Kff^{(s)} - \Kfu^{(s)} \Kuu^{-1} \Kuf^{(s)} + \sigma_w^2\mathbf{I},
\end{align*}
where $\Kfu^{(s)}=k_w(\Cvec_\wvec^{(s)}, \Cvec_{\uvec})$, $\Kuu=k_w(\Cvec_{\uvec}, \Cvec_{\uvec})$, $\Kff^{(s)}=k_w(\Cvec_\wvec^{(s)}, \Cvec_\wvec^{(s)})$. At this point, we can either (i) sample $\uvec$ from $q(\uvec)$, or (ii) integrate $\uvec$ out analytically. Opting for the second approach gives $q(\wvec | \zvec_s) = \norm(\wvec; \Avec^{(s)} \muvec_\uvec, \Bvec^{(s)} + \Avec^{(s)} \Sigma_\uvec \Avec^{\intercal, (s)})$, the former just omits the second covariance term and uses a sample $\uvec^{s}$ for the predictive mean instead of $\muvec_\uvec$.

\begin{figure}[h]
\centering
\includegraphics[width=0.6\textwidth]{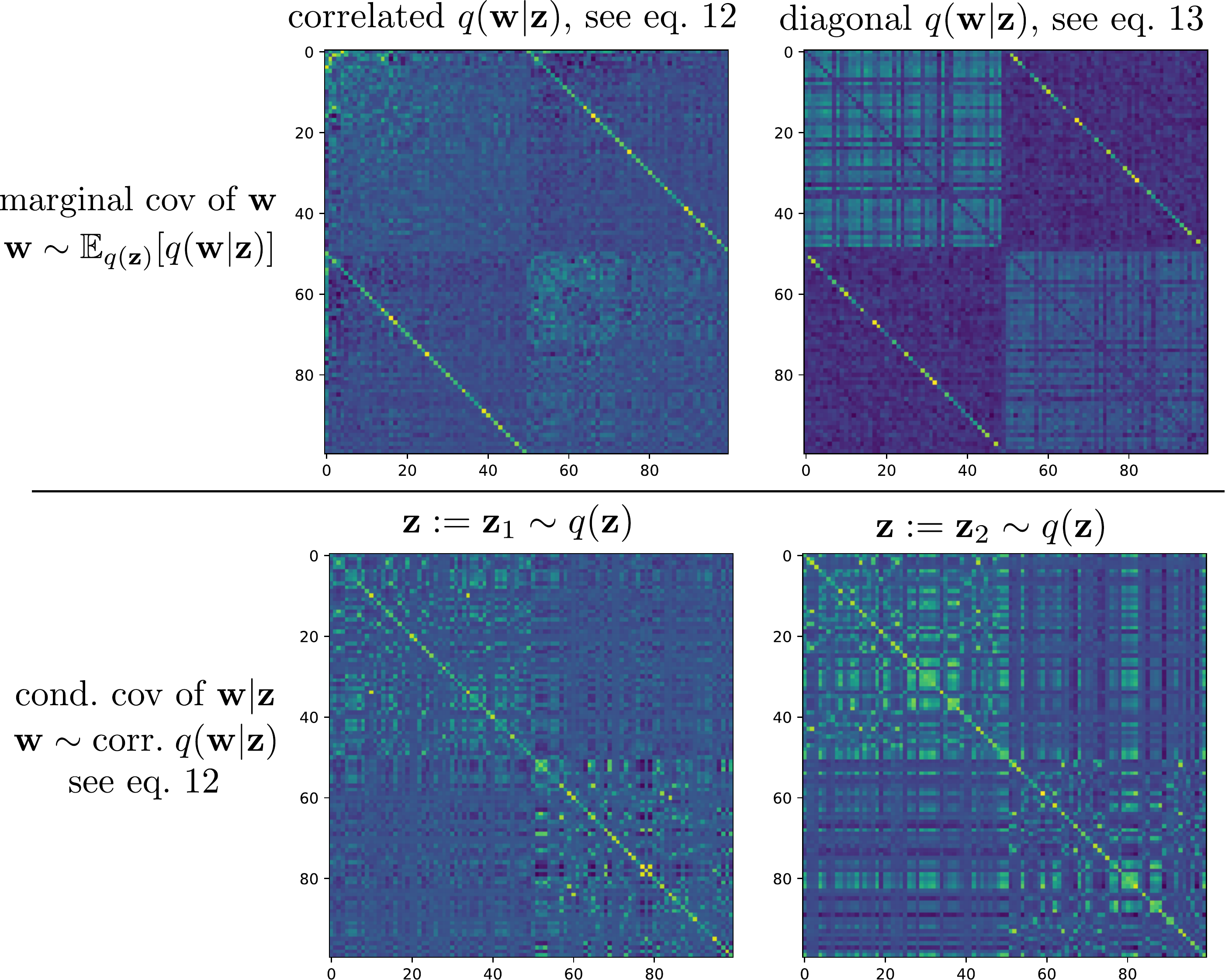}
\caption{Marginal and conditional covariance structures over weights in a 1x50x1 BNN. Sampling from the posterior of the hierarchical model reveals that even a diagonal GP approximation can capture off-diagonal correlations induced through unit correlations. Also note the off-diagonal bands in the marginal plots above, which indicate the correlation structures induced by the latent variables of the hidden units connecting the layers.We remove the diagonal in the marginal plots for clarity.}
\label{fig:weight_cov}
\end{figure}

In contrast to GP regression and classification in which the likelihood term is factorized point-wise w.r.t.~the parameters and thus their expectations only involve a low dimensional integral, we have to integrate out $\wvec$ which for GPs entails inversion of the $|\wvec| \times |\wvec|$ matrix {\bf K} (which is $\Bvec^{(s)}$ when we don't sample $\uvec$ or the full term above). This is feasible for small neural networks with up to a few thousand weights, but becomes intractable for more general architectures. In order to scale to larger networks, we introduce a diagonal approximation, which given a sample $\uvec^{s}$ looks as follows, $\hat{q}(\wvec| \zvec_s) = \norm(\wvec; \Avec^{(s)} \uvec^{s}, \diag(\Bvec^{(s)}))$.
Whilst the diagonal approximation above might look poor at first glance, it is conditioned on a sample of the latent variables $\zvec_s$ and thus the weights' correlations induced by the hierarchical unit-structure are retained after integrating out $\zvec$. Such correlations are illustrated in \cref{fig:weight_cov}, showing the marginal and conditional covariance structures for the weights of a small neural network, separated into diagonal and full covariance models.
We also provide a qualitative and quantitative analysis of performance of different approximations to $q(\wvec | \zvec_s)$ in the appendix, including the diagonal approximation presented here, and show that not only is this approximation fast but also that it performs competitively with full covariance models.
Finally, the expected log-likelihood $\mathcal{F}_s$ is approximated by $\mathcal{F}_{s} \approx \frac{1}{J}\sum \limits_{j=1}^{J}\log p(\yvec | \wvec_{j}, \xvec)$ with samples $\wvec_j \sim \hat{q}(\wvec| \zvec_s)$ \footnote{We can also use the {\it local reparameterization trick} \citep{kingma2015variational} to reduce variance.}. The final lower bound is then optimized to obtain the variational parameterers of $q(\uvec)$, $q(\zvec)$, and estimates for the noise in the meta-GP model, the kernel hyper-parameters and the inducing inputs.  
%


\begin{figure*}[!ht]
    \centering
    \includegraphics[width=1.0\textwidth]{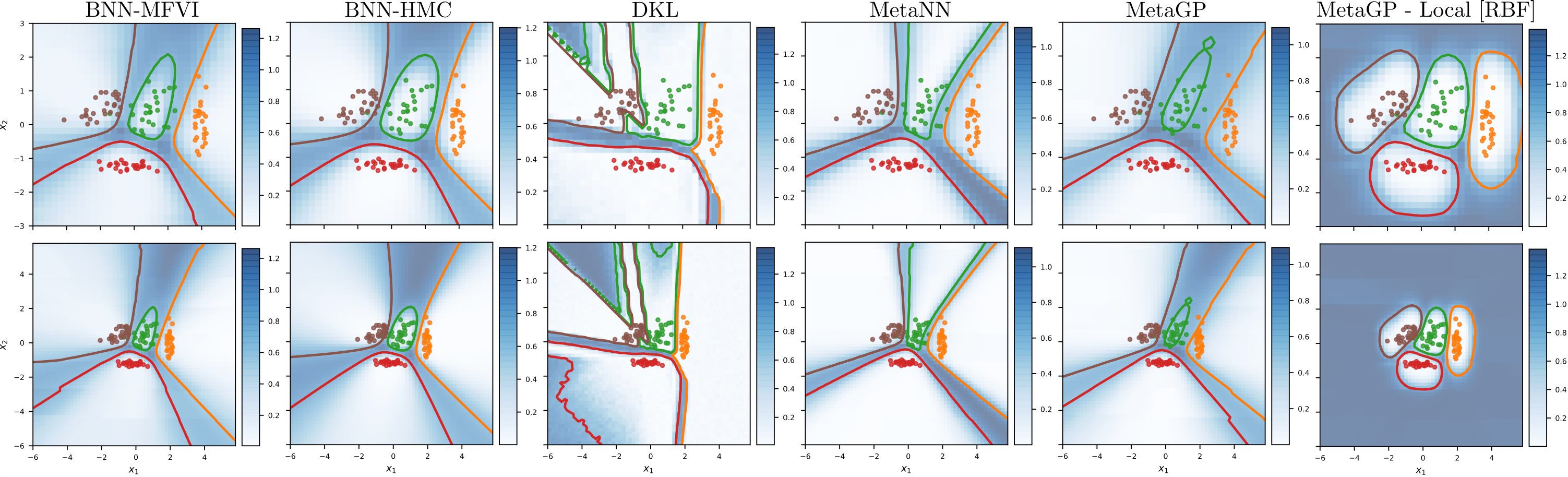}
    \caption{Predictive performance of various methods on a four-way classification problem. We compare the proposed approaches (MetaGP, MetaGP with an input-dependent RBF kernel and periodic kernel) to BNN with MFVI and HMC, DKL and MetaNN. Best viewed in colour. The background color shows the entropy of the predictive distribution. The contours show the 0.7 equiprobability contours. The bottom plots are the zoom-out version of the corresponding top plots, showing the predictive entropy further from the training points.}
    \label{fig:toy_cla}
\end{figure*}

\subsection{Inference for the local model}
\label{sec:inference_local}
The main difference in the local model is the dependence of weights on inputs.
To handle inducing point kernels over both weight codes and inputs, we introduce inducing inputs $\tilde{\epsvec} = \{\tilde{\epsilon}_m\}_{m=1}^{M}$ where $\tilde{\epsilon} \in \mathcal{R}^{D_{\mathrm{aux}}}$ for $k_{\mathrm{aux}}$. We then concatenate the dimensions of $\Cvec_{\uvec}$ in \cref{sec:inference_global} with the new inducing inputs to form the new inputs $\tilde{\Cvec}_{\uvec} = [\Cvec_{\uvec}; \tilde{\epsvec}]$. The set of inputs $\tilde{\Cvec}_{\uvec}$ now have dimensions $\tilde{\cvec}_{u}  \in \mathcal{R}^{2D_z+ D_{\mathrm{aux}}}$.
The fully instantiated covariance matrix ${\bf K}_{\mathrm{local}} = K_w \otimes K_{\mathrm{aux}} $ would take the shape $|\wvec| \times |\wvec| \times N \times N$. As this kernel has Kronecker structure one could now consider using inference techniques such as in~\citep{flaxman2015fast}. However, the tractability of the global kernel remains an issue even in this case. As such, we elect to inherit the diagonal approximation from~\cref{sec:inference_global} and apply it to the joint kernel, yielding an object of dimension $|\wvec| \times N$. The lower bound computation in \cref{sec:inference_global} can thus be reused but with $\Avec^{(n,s)}$ and $\Bvec^{(n,s)}$ being input-dependent\footnote{Specifically, $\Avec^{(n,s)} = \Kfu^{(n,s)} \Kuu^{-1},
\Bvec_n^{(n,s)} = \Kff^{(n,s)} - \Kfu^{(n,s)} \Kuu^{-1} \Kuf^{(n,s)} + \sigma_w^2\mathbf{I},$
where $\Kfu^{(n,s)}=k_w(\Cvec_\wvec^{(s)}, \Cvec_{\uvec}) \otimes k_{\mathrm{aux}}(\xvec_n, \tilde{\epsvec})$, $\Kuu=k_w(\Cvec_\uvec, \Cvec_{\uvec}) \cdot k_{\mathrm{aux}}(\tilde{\epsvec}, \tilde{\epsvec})$, $\Kff^{(n,s)}=k_w(\Cvec_\wvec^{(s)}, \Cvec_\wvec^{(s)}) \otimes k_{\mathrm{aux}}(\xvec_n, \xvec_n)$.}. We can handle large datasets by using inducing point kernels, which permit inference using minibatches. Another difference is the potential existence of the mapping $\Vvec$ in the model, which we tackle by introducing a variational distribution $q(\Vvec) = \norm(\Vvec; \muvec_\Vvec, \diag(\sigmasvec_\Vvec))$. We can estimate the evidence lower bound by also drawing unbiased samples from this and jointly optimizing its parameters with the rest of the variational parameters. The overall computational complexity with data-subsampling in this section and \cref{sec:inference_global} is $\mathcal{O}(M^3 + |\wvec|M^2)$.

\section{Experiments}
\label{sec:exp}
In this section, we evaluate the proposed priors and inference scheme on several regression and classification datasets. These were implemented using PyTorch \citep{paszke2017automatic} and the code will be available upon acceptance. Additional results are included in the appendices. We use $M=50$ inducing points for all experiments in this section. All experiments were run on a Macbook pro.

\begin{figure*}[!ht]
    \centering
    \includegraphics[width=1.0\textwidth]{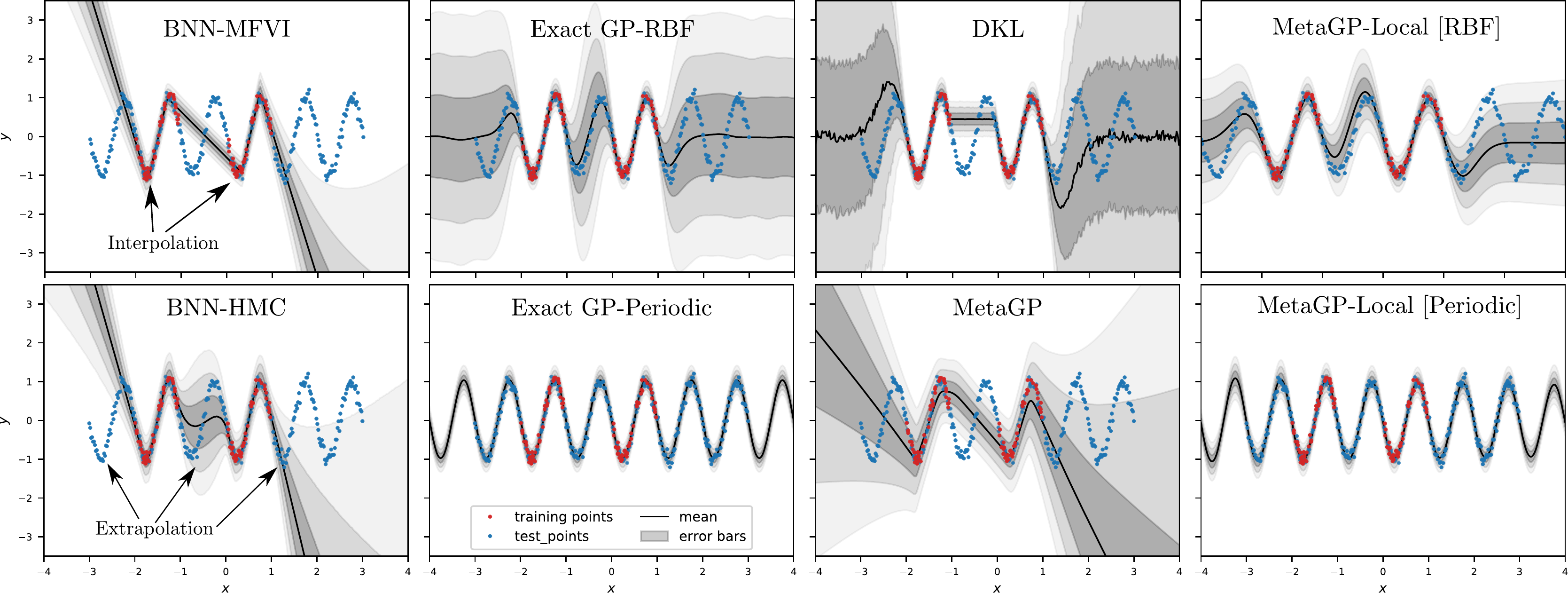}
    \caption{Predictive performance of various methods on a sinusoidal dataset. We also provide a quantitative comparison in \cref{tab:sinusoid_results}.}
    \label{fig:sinusoid}
\end{figure*}

\subsection{Synthetic classification example}
We first illustrate the performance of the proposed model on a classification example. We generate a dataset of 100 data points and four classes, and use a BNN with one hidden layer of 50 hidden units with ReLU non-linearities, and two dimensional latent variables {\bf z}. \Cref{fig:toy_cla} shows the predictive performance of the proposed priors and various alternatives, including BNN (with unit Normal priors on weights) with mean field Gaussian variational approximation (MFVI) \citep{blundell2015weight} and Hamiltonian Monte Carlo (HMC) \citep{neal1992bayesian}, variational deep kernel learning (DKL) \cite{wilson2016stochastic} and MetaNN \citep{karaletsos2018probabilistic}. We highlight that {\it MetaGP-local} with RBF kernel gives uncertainty estimates that are reminiscent to that of a GP model in that the predictions express {\it ``I don't know"} away from the training data, despite being a neural network under the hood. Following \cite{bradshaw2017adversarial}, we also show the uncertainty for data further from the training instances. {\it MetaGP-local}(RBF) remains uncertain, as expected, for these points while MFVI and DKL produce arguably overconfident predictions.

\subsection{Inductive Biases For Neural Networks With Input-Dependent Kernels}
We explore the utility of the input-dependent prior towards modeling inductive biases for neural networks and evaluate predictive performance on a regression example. In particular, we generate 100 training points from a synthetic sinusoidal function and create two test sets that contain in-sample inputs and out-of-sample inputs, respectively. We test an array of models and inference methods, including BNN (with unit Normal priors on weights) with MFVI and HMC, GPs with diverse kernel functions, DKL, MetaGP and local-MetaGP with input dependence given the same kernels as the GPs. We use RBF and periodic kernels \citep{mackay1998introduction} for weight modulation and the pure GP in this example. \Cref{fig:sinusoid} summarizes the results. Note that the periodic kernel allows the BNN to discover and encode periodicity in its weights, leading to long-range confident predictions compared to that of the RBF kernel and significantly better extrapolation than BNNs with independent weight priors can obtain given the amount of training data, even when running HMC instead of VI.

We evaluate the quantitative utility of input-dependence and inductive biases on two test sets that contain in-sample inputs (between the training data) and out-of-sample inputs (outside the training range), respectively. We report the performance of all methods in \cref{tab:sinusoid_results}. The performance is measured by the root mean squared error (RMSE) and the negative log-likelihood (NLL) on the test set, and we explicitly evaluate separately for {\it extrapolation} and {\it interpolation}. In this example, the local MetaGP model is comparable to GP regression with a periodic kernel and superior to other methods, demonstrating good RMSE and NLL on both in-distribution and out-of-distribution examples.

\begin{table}[!ht]
\vspace{-10pt}
\small
\centering
\caption{Average test error and negative log-likelihood for the sinusoid example in \cref{fig:sinusoid}, averaged over five runs. Lower is better.}
\label{tab:sinusoid_results}
\begin{tabular}{c|cc|cc}
\hline
& \multicolumn{2}{c|}{Interpolation} & \multicolumn{2}{c}{Extrapolation} \\
\hline
Method & RMSE & NLL & RMSE & NLL \\
\hline
\hline
BNN-MFVI&0.17&-0.04&3.51&88.12\\
BNN-HMC&0.12&-0.69&4.34&10.98\\
Exact GP-RBF&\textbf{0.11} &\textbf{-0.81} &0.55&0.75\\
Exact GP-Periodic&\textbf{0.11} &-0.80&\textbf{0.11} &\textbf{-0.83} \\
DKL&0.12&-0.72&0.76&3.26\\
MetaGP&0.24&0.08&2.59&5.86\\
MetaGP-Local[RBF]&\textbf{0.11} &-0.80&0.74&1.50\\
MetaGP-Local[Periodic]&\textbf{0.11} &-0.76&0.12&-0.69\\
\hline
\end{tabular}%
\vspace{-5pt}
\end{table}

\begin{figure*}[!ht]
    \centering
    \includegraphics[width=1.00\textwidth]{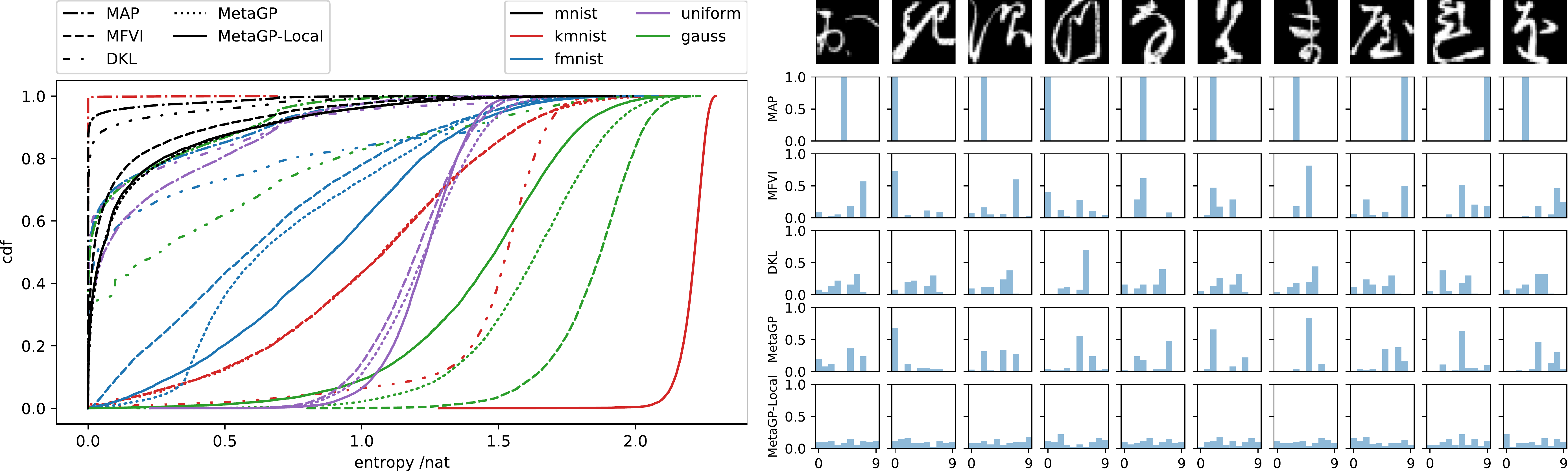}
    \caption{The CDFs of predictive entropies on in-distribution and out-of-distribution test sets for various methods [Left] and the predictive class probability for representative samples from out-of-distribution test sets [Right].}
    \label{fig:mnist_ood}
\end{figure*}

\subsection{Input Dependent Neural Networks For Uncertainty Quantification}
Motivated by the performance of the proposed {\it MetaGP-local} model in the synthetic examples in \Cref{fig:sinusoid}, this section tests the ability of this model class to produce calibrated predictive uncertainty to out-of-distribution samples. That is, for test samples that do not come from the same training distribution, a robust and well-calibrated model should produce uncertain predictive distribution and thus high predictive entropy. Such a model could find applications in safety-critical tasks or in an area where detecting unfamiliar inputs is crucial such as active learning or reinforcement learning. In this experiment, we train a BNN classifier with one hidden layer of 100 rectified linear units on the MNIST dataset, with {\it MetaGP-local}-RBF only applied to the last layer of the network. The dimensions of the latent variables and the auxiliary inputs are both 2, with auxiliary inputs given by transforming MNIST images using a jointly learned linear projection $\Vvec$. After training on MNIST, we compute the entropy of the predictions on various test sets, including notMNIST, fashionMNIST, Kuzushiji-MNIST, and uniform and Gaussian noise inputs. Following \citep{lakshminarayanan2017simple, louizos2017multiplicative}, the CDFs of the predictive entropies for various methods are shown in \cref{fig:mnist_ood}. A calibrated classifier should give a CDF that bends towards the top-left corner of the plot for in-distribution examples and, vice versa, towards the bottom-right corner of the plot for out-of-distribution inputs. In most out-of-distribution sets considered, except Gaussian random noise, {\it MetaGP} and {\it MetaGP-local} demonstrate superior performance to all comparators, including DKL. Notably, MAP estimation, often deployed in practice, tends to give wildly poor uncertainty estimates on out-of-distribution samples. We illustrate this behaviour and that of other methods on representative inputs of the Kuzushiji-MNIST dataset in \Cref{fig:mnist_ood} and on MNIST digits in the appendix.

\subsection{Active learning}
\label{sec:active_learning}
\begin{figure}[!ht]
    \centering
    \includegraphics[width=\columnwidth]{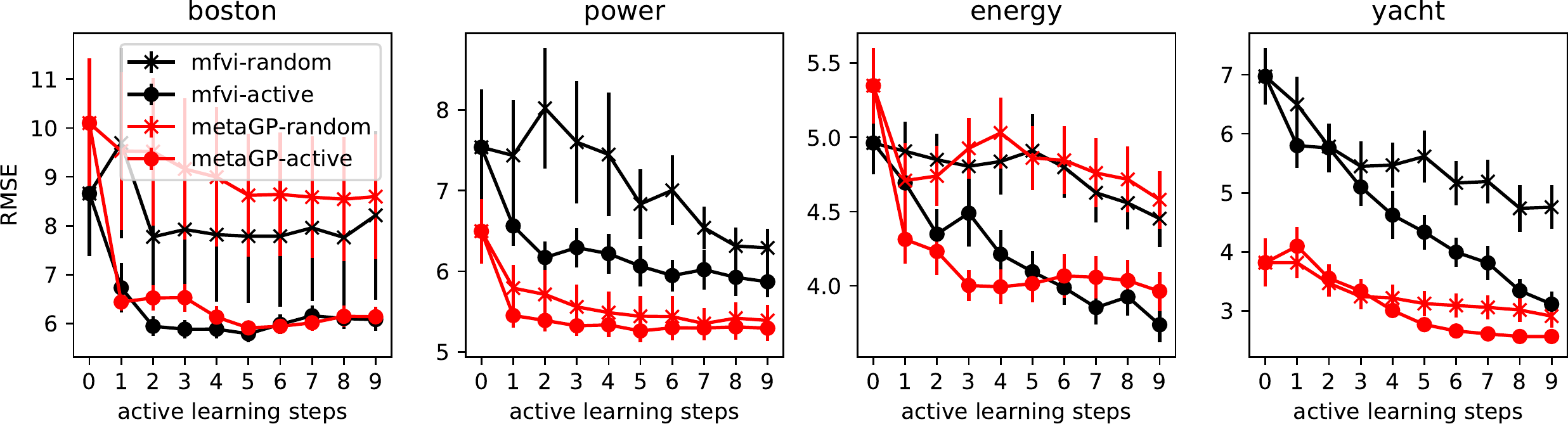}
    \caption{Active learning with BNNs using mean-field Gaussian variational inference [MFVI] and a meta-GP hierarchical prior [MetaGP] on several UCI regression datasets. Each trace shows the root mean squared error (RMSE) averaged across 40 runs.}
    \label{fig:active_uci_rmse}
\end{figure}
We next stress-test the performance of the proposed model in a pool-based active learning setting for real-valued regression, where limited training data is provided initially and the target is to sequentially select points from a pool set to add to the training set. The criterion to select the next best point from the pool set is based on the entropy of the predictive distribution, i.e.~we pick one with the highest entropy. Note that this selection procedure can be interpreted as selecting points that maximally reduce the posterior entropy of the network parameters \cite{houlsby2011bayesian}. Four UCI regression datasets were considered, where each with 40 random train/test/pool splits. For each split, the initial train set has 20 data points, the test set has 100 data points, and the remaining points are used for the pool set, similar to the active learning set-up in \cite{hernandez-lobato2015prob}. We compare the performance of the proposed model and inference scheme to that of Gaussian mean-field variational inference and show the average results in \Cref{fig:active_uci_rmse}. Across all runs, we observe that active learning is superior to random selection and more crucially using the proposed model and inference scheme seems to yield comparable or better predictive errors with a similar number of queries.

\section{Related work}
\label{sec:related}
There is a long history of research on developing (approximate) Bayesian inference methods for BNNs, i.e.~in \citep{neal1992bayesian,neal2012bayesian,ghahramani2016history}. 
Our work differs in that the model employs a hierarchical prior, and inference is done in a lower-dimensional latent space instead of the weight space. The variational approximation is chosen such that the marginal distribution over the weights is non-Gaussian and the correlations between weights are retained, in contrast to the popular mean-field Gaussian approximation.
Imposing structure over the weights with a carefully chosen prior has been observed to improve predictive performance \citep{ghosh2018structured,neal2012bayesian,blundell2015weight}, but it has remained elusive how to express prior knowledge or handle interpolation or extrapolation in such models.
Modern deep Bayesian learning approaches often involve fusing neural networks and GPs, such as in deep kernel learning \citep{wilson2016stochastic}, which layers a GP on top of a neural network feature extractor.
Another notable example is~\citep{pearce2019expressive}, which blends kernels and activation functions to induce
desired properties through architectural choices, but is not expressing these assumptions as a weight prior. The functional regularization approach introduced in \citep{sun2019functional} shares some of the motivations with our paper, but implements it very differently by explicitly instantiating a GP and performing a complex training scheme to learn neural networks that match that GP. Asymptotically, they match the GP, while in our model (i) the properties we care about are already built into the weight prior allowing direct training on a dataset without the involved minimax approach, and (ii) our posterior can depart from that restrictive prior as it fundamentally only guides a weight based model, i.e. by learning posterior kernel parameters for $k_{\mathrm{aux}}$ to eliminate its influence on $k_{\mathrm{local}}$ (such as wide lengthscales).\\
Another related theme is hyper-networks, the core idea of which is to generate network parameters using another network \citep[see e.g.][]{ha2016hypernetworks,stanley2009hypercube}. Our model resembles a GP-LVM~\citep{lawrence2004gaussian} hyper-GP, with a key structural assumption of node latent variables as introduced in \citep{karaletsos2018probabilistic} to enable compact prediction per weight instead of per weight tensor.


\section{Summary}
\label{sec:summary}
We proposed a GP-based hierarchical prior over neural network weights, and a modification that permits input-dependent weight priors, along with an effective approximate inference strategy. We demonstrated utility of these models for interpolation, extrapolation, uncertainty quantification and active learning benchmarks, outperforming strong baselines. We plan to evaluate the performance of the model on more challenging decision making tasks.

\bibliography{gp_metarep_arxiv.bib}
\bibliographystyle{abbrvnat}

\appendix

\section{Additional Background on Bayesian neural networks and variational inference}
\label{sec:bayesiannn}
Consider a training set comprising of $N$ input-output pairs, $\Dcal = \{\xvec_n, y_n\}_{n=1}^N$, and a neural network parameterized by weights and biases, $\wvec$, that describes the distribution over an output $y_n$ given an input $\xvec_n$, $p(y_n | \wvec, \xvec_n)$. We follow a Bayesian approach by placing a prior distribution over the network parameters, $p(\wvec)$, and obtaining the posterior distribution $p(\wvec | \Dcal)$, which involves calulation of the marginal likelihood $p(\Dcal) = \int \dd \wvec p(\wvec) p(\Dcal | \wvec)$.
However, obtaining $p(\wvec | \Dcal)$ and $p(\Dcal)$ exactly is intractable when $N$ is large or when the network is large and as such, approximation methods are often required. In particular, mean-field Gaussian variational inference (MFVI) has recently become a method of choice for approximate inference for Bayesian neural networks due to its simplicity and the recently popularized {\it reparameterization trick} \citep{salimans2013fixed,kingma2013auto,titsias2014doubly,blundell2015weight}. MFVI sidesteps the intractability by positing a diagonal Gaussian approximation $q(\wvec) = \norm(\wvec; \muvec, \diag(\sigmasvec))$ and optimising an approximate lower bound to the marginal likelihood $\Lcal_{\mathrm{MFVI}}(q(\wvec))\approx - \mathrm{KL} [q(\wvec) || p(\wvec)] + \frac{1}{K} \sum_{k=1}^{K} \sum_{n=1}^{N} \log p(y_n | \wvec_k, \xvec_n)$,
where $\wvec_k = \muvec + \sigmavec \odot \epsilon_k$ and $\epsilon_k \sim \norm(\mathbf{0}, \mathbf{I})$, i.e.~$\wvec_k$ is a sample from $q(\wvec)$. 
Note that the mean-field variational Gaussian approximation with a standard normal prior, presented in is often outperformed by point estimation in certain settings \citep{trippe2018overpruning}.
Despite being practical and able to give reasonable uncertainty estimates, improving MFVI is still an active research area, and the main focuses of which are (i) improving the reparameterization gradient estimator to enable faster convergence \citep{miller2017reducing,wu2018fixing}, (ii) replacing the typical standard Normal prior, $p(\wvec) = \norm(\wvec; \mathbf{0}, \mathbf{I})$ by a structured prior that better models the structures present in the weight a-priori \citep{ghosh2018structured,neal2012bayesian,blundell2015weight}, and (iii) using structured variational approximations that can potentially capture weight correlations in the posterior \citep{louizos2016structured,zhang2017noisy}. This paper builds on the two latter themes and proposes a hierarchical model for the prior and a structured variational scheme that explicitly model and infer weight structures.

\section{Extra experimental results}

\subsection{An empirical evaluation of various approximations for $q(\wvec | \zvec_k, \Vvec_k, {\bf x})$}

In this section, we analyze the impact of different approximations to the covariance matrix of $q(\wvec | \zvec_k, \Vvec_k, {\bf x})$:
\begin{align}
    q(\wvec | \zvec_k, \Vvec_k, {\bf x}) = \norm(\wvec; \Avec^{(k)} \muvec_u, \Bvec^{(k)} + \Avec^{(k)} \Sigma_\uvec \Avec^{\intercal, (k)}). \nonumber
\end{align}
If we use the exact, fully correlated Gaussian distribution above, it is necessary to sample from this distribution to evaluate the lower bound. This step costs $\mathcal{O}(W^3)$ where $W$ is the number of parameters in the network.

The complexity can be greatly improved by making a diagonal approximation to $\Bvec^{(k)}$ as follows, 
\begin{align}
  \hat{q}_{\mathrm{FITC}}(\wvec| \zvec_k, \Vvec_k, {\bf x}) = \norm(\wvec; \Avec^{(k)} \muvec_u, \diag(\Bvec^{(k)}) + \Avec^{(k)} \Sigma_\uvec \Avec^{\intercal, (k)}). \nonumber
\end{align}
Sampling from this distribution can be done in $\mathcal{O}(WM^2)$ where M is the number of pseudo-points.

This can be further approximated by assuming a diagonal covariance matrix,
\begin{align}
  \hat{q}_{\mathrm{diag}}(\wvec| \zvec_k, \Vvec_k, {\bf x}) = \norm(\wvec; \Avec^{(k)} \muvec_u, \diag(\Bvec^{(k)} + \Avec^{(k)} \Sigma_\uvec \Avec^{\intercal, (k)})). \nonumber
\end{align}
The variational bound can then be evaluated by drawing samples from $\hat{q}_{\mathrm{diag}}$ as in the above approximation, or by drawing activity samples by employing the local reparameterization trick \citep{kingma2015variational}.

\begin{figure}[!ht]
    \centeringœ
    \includegraphics[width=\textwidth]{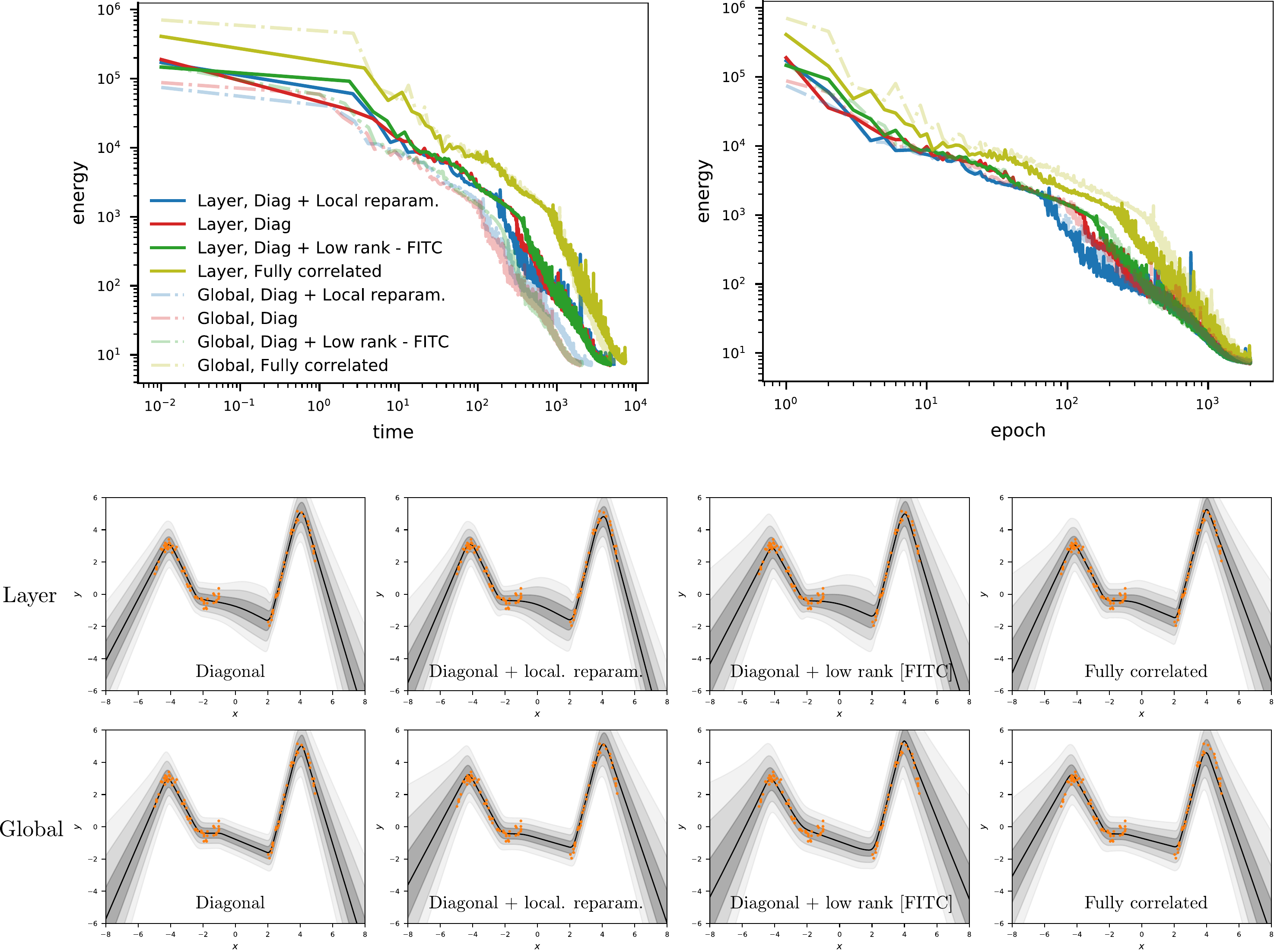}
    \caption{An evaluation of the covariance matrix approximations in a toy regression example. Top: objective function during training vs epoch/time. Bottom: Predictions after training using one of the approximations discussed in the text. Global: there is one GP for all weights in the network. Layer: there are multiple GPs, one for each weight layer in the network. Note that we are not using the auxiliary kernel here. Best viewed in colour.}
    \label{fig:diag_corr_xsin}
\end{figure}

\begin{figure}[!ht]
    \centering
    \includegraphics[width=\textwidth]{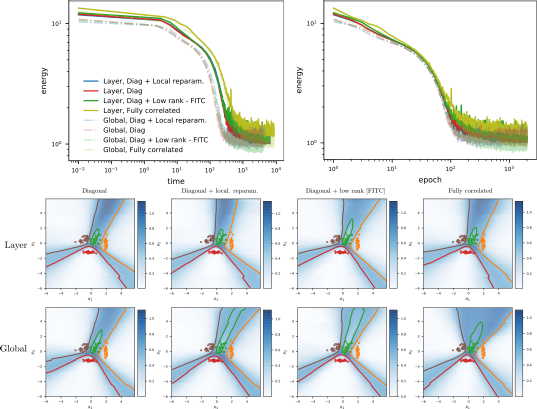}
    \caption{An evaluation of the covariance matrix approximations in a toy classification example. Top: objective function during training vs epoch/time. Bottom: Predictions after training using one of the approximations discussed in the text. Global: there is one GP for all weights in the network. Layer: there are multiple GPs, one for each weight layer in the network. Note that we are not using the auxiliary kernel here. Best viewed in colour.}
    \label{fig:diag_corr_four}
\end{figure}

We evaluate the performance of using the exact and approximate conditional distributions above in a range of toy regression and classification, and show representative results in \cref{fig:diag_corr_xsin,fig:diag_corr_four}. We note that the diagonal approximation is fast and gives qualitatively similar performance compared to more structured approximation or the exact case, in both cases where there is a single GP for all weights in the network and there is multiple GPs, one for each weight layer in the network.

\clearpage

\subsection{Results on a synthetic regression example}
\begin{figure*}[!ht]
    \centering
    \includegraphics[width=1.0\textwidth]{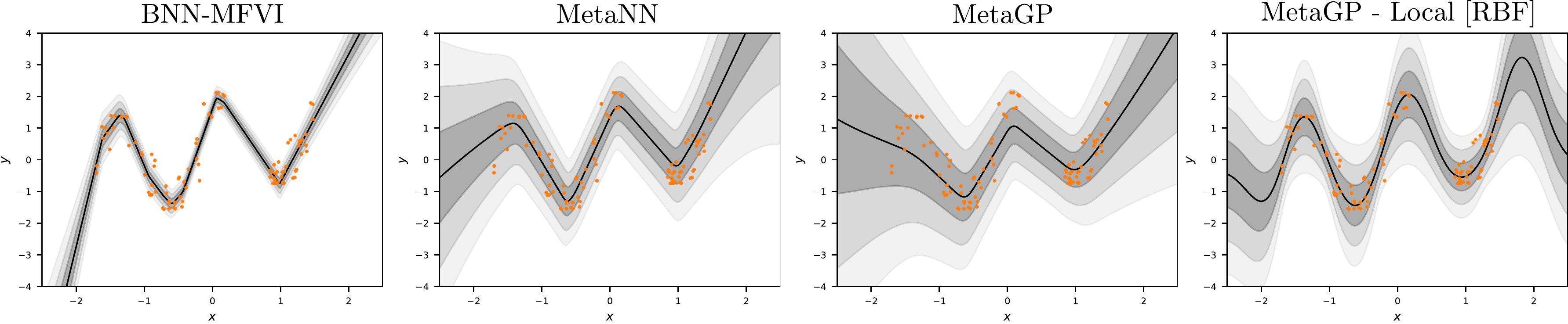}
    \caption{Predictive performance of various methods on a 1D test function. We compare the proposed approaches (MetaGP and MetaGP with an input-dependent kernel) to BNN-MFVI and MetaNN. Best viewed in colour.}
    \label{fig:toy_reg}
\end{figure*}

In this section, we demonstrated the performance of the proposed priors on a 1D test function, as used in \citep{louizos2019fnp}. We compare to BNN with independent Gaussian priors and a mean-field Gaussian variational approximation, and MetaNN \citep{karaletsos2018probabilistic}. The training points, and predictive mean and error bars are shown in \cref{fig:toy_reg}.

\subsection{Robustness in various data regimes for a toy regression problem}
\begin{figure}[!ht]
    \centering
    \includegraphics[width=\textwidth]{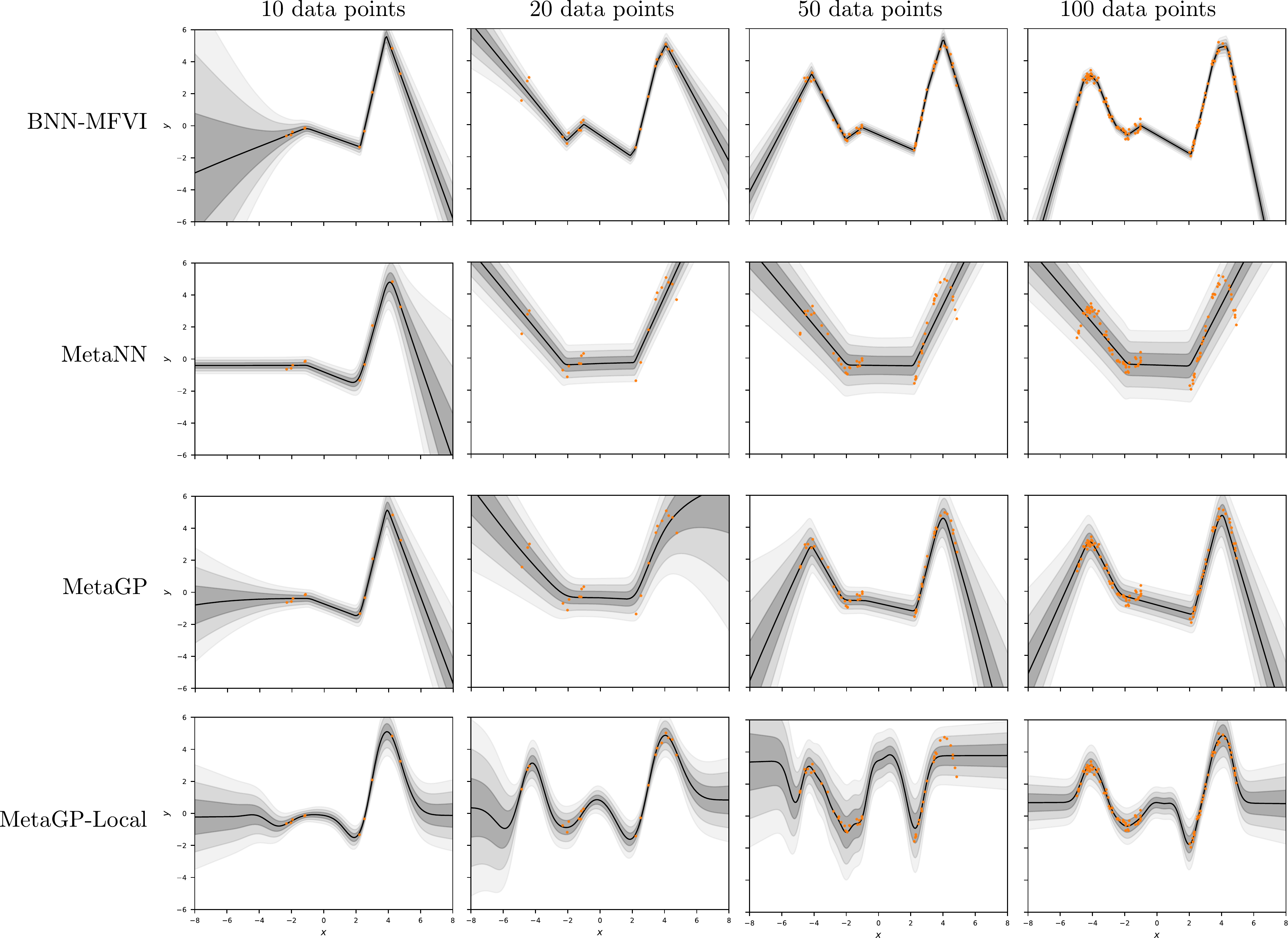}
    \caption{Performance of mean-field variational inference, MetaNN with variational inference and MetaGP with variational inference on a toy regression problem with various number of training points. Best viewed in colour.}
    \label{fig:xsin_regime}
\end{figure}

In this experiment, we evaluate the qualitative performance of various methods, including MFVI, MetaNN, MetaGP and MetaGP with local, input-dependent kernel, on a toy regression problem, in different data regimes. In particular, we considers 10, 20, and 50 training points respectively, and plot the predictions in \cref{fig:xsin_regime}. MetaGP demonstrates consistent performance across all data regimes, comparable to that of MetaNN. The input-dependent kernel helps the performance further in the out-of-distribution area.

\clearpage

\subsection{Robustness of MetaGP with network architectures}
\begin{figure}[!ht]
    \centering
    \includegraphics[width=\textwidth]{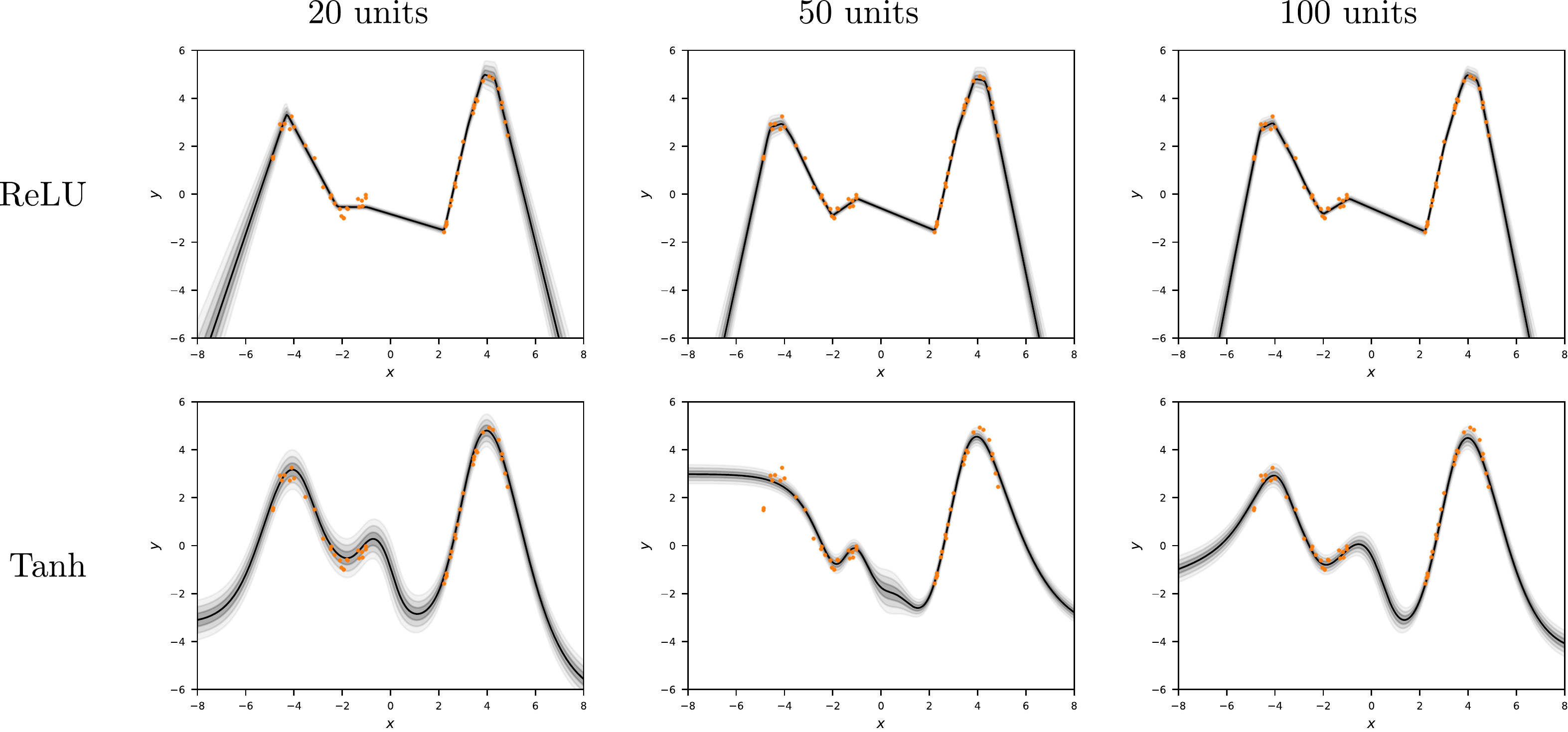}
    \caption{Performance of MetaGP on a toy regression problem, with various numbers of hidden units and different activation functions. Best viewed in colour.}
    \label{fig:gp_metarep_arch}
\end{figure}

In this experiment, we compare the performance of MetaGP for various numbers of hidden units (20, 50 and 100) and two activation functions (Tanh and ReLU) on a toy regression problem. The observation noise is fixed in this experiment. We observe that the performance of the models is in general consistent across different activation functions and numbers of hidden units. We show the results in \cref{fig:gp_metarep_arch}.

\clearpage

\subsection{Effect of input-dependent kernels}

\begin{figure}[!ht]
    \centering
    \includegraphics[width=0.6\textwidth]{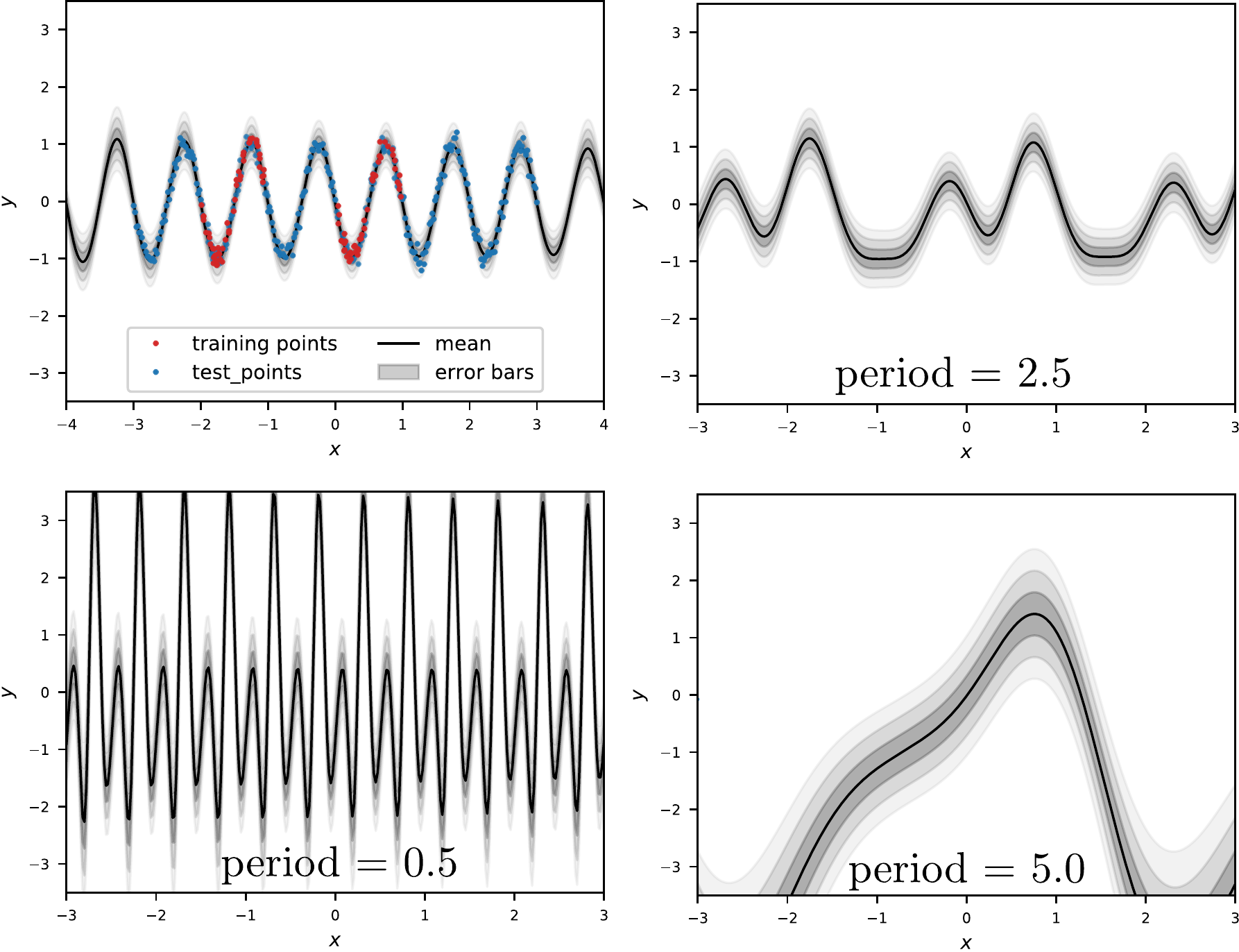}
    \caption{We first train a model with an input-dependent kernel on a sinusoid data set (top left) and then vary the period hyperparameter of the input-dependent kernel whilst keeping other hyperparameters and variational parametes fixed (others). Best viewed in colour.}
    \label{fig:sin_period}
\end{figure}

To understand the impact of the auxiliary kernel to the prediction, we use a model trained on the sinusoid dataset, as shown in the main text, and vary the period hyper-parameter in the kernel whilst keeping other hyper-parameters and variational parameters fixed. The predictions for a few hyperparameters are shown in \cref{fig:sin_period}. We note the variation/period in the data is captured by weight modulation, governed by the input-dependent kernel. Changing the period hyperparameter affects how fast or slow the weights are changing wrt the input.

\clearpage

\subsection{MNIST experiment: full figures}

\begin{figure}[!ht]
    \centering
    \includegraphics[width=\textwidth]{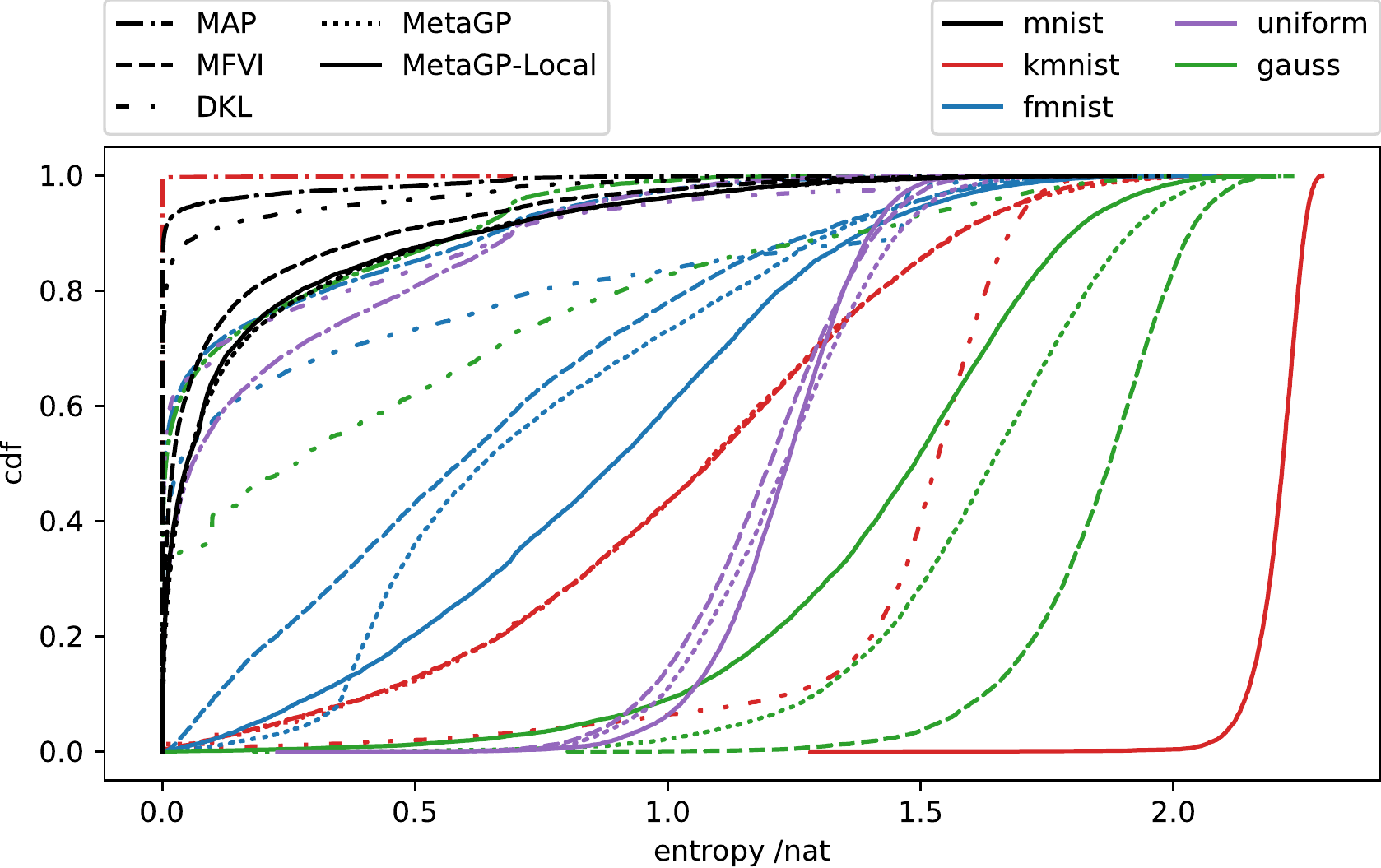}
    \caption{Full results of the MNIST out-of-distribution uncertainty experiment. Best viewed in colour.}
    \label{fig:mnist_entropy_all}
\end{figure}

\begin{figure}[!ht]
    \centering
    \includegraphics[width=\textwidth]{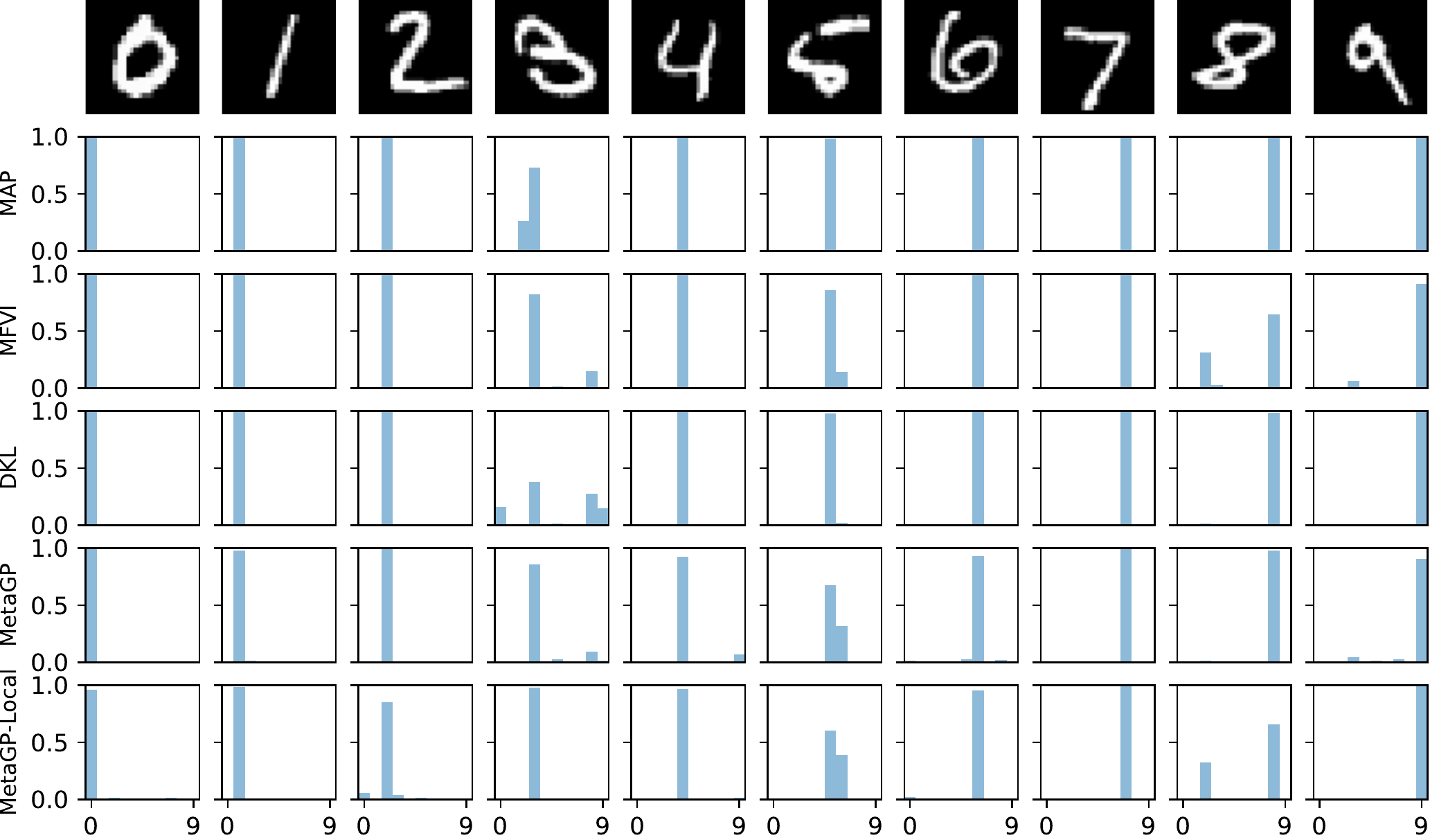}
    \caption{Predictive distribution for representative MNIST test examples by various methods. Best viewed in colour.}
    \label{fig:mnist_mnist}
\end{figure}

\begin{figure}[!ht]
    \centering
    \includegraphics[width=\textwidth]{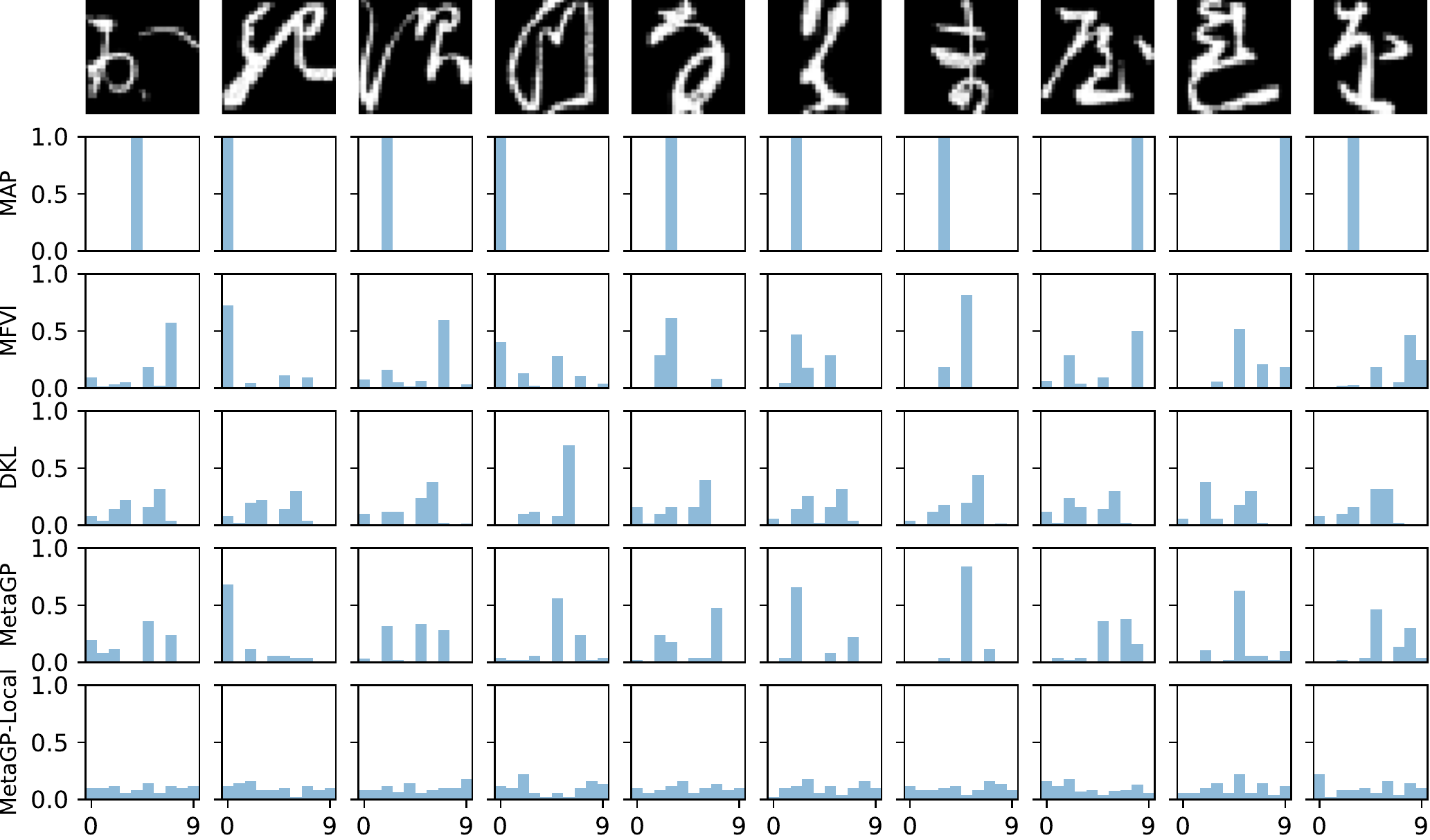}
    \caption{Predictive distribution for representative KMNIST test examples by various methods. Best viewed in colour.}
    \label{fig:mnist_kmnist}
\end{figure}

In this section, we include the full figures of the MNIST out-of-distribution uncertainty experiment, as well as additional results using deep kernel learning \citep{wilson2016deep}. In particular, we employ the same network architecture with the last layer being replaced by multiple independent GPs, one for each class (output dimension). As exact inference is intractable, variational inference based on inducing points is employed -- we used 50 inducing points for each output. The full results of all models/methods considered are shown in \cref{fig:mnist_entropy_all}. For clarify, the results of deep kernel learning and MetaGP are shown in \cref{fig:mnist_entropy_3}. MetaGP with the input-dependent kernel shows good performance, outperforming deep kernel learning in all cases. In addition, we include the full figures for the predictive distributions on representative test examples in \cref{fig:mnist_mnist,fig:mnist_kmnist}. 

\clearpage
\subsection{A toy active learning problem}
\begin{figure}[!ht]
    \centering
    \includegraphics[width=\textwidth]{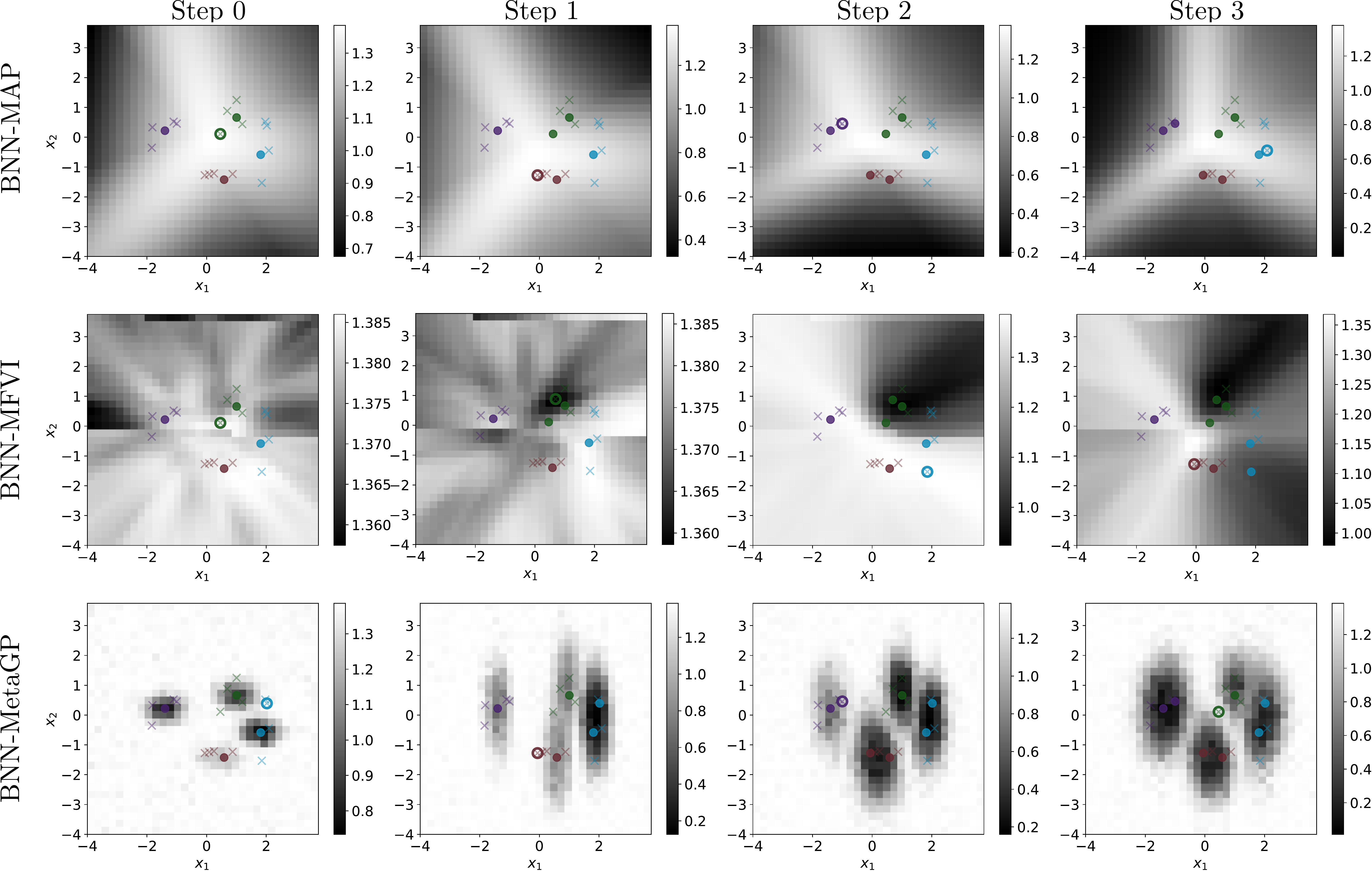}
    \caption{Active learning with BNNs using maximum a posteriori estimation [BNN-MAP], mean-field Gaussian variational inference [BNN-MFVI] and a meta-GP hierarchical prior [BNN-MetaGP] on a toy multi-class classification task. For each plot, the filled circle markers are the current training points, with different colours illustrating different classes. The shaded crosses are the examples in the pool set, one of which we wish to pick and evaluate to be included in the training set. The unfilled circle markers are the examples from the pool set selected at a step. The objective function for selecting points from the pool set is the entropy of the predictive probability. Best viewed in colour.}
    \label{fig:active_toy}
\end{figure}

In this section, we provide a visualisation of the predictive performance of different methods in an active learning setting. Please see \cref{fig:active_toy} and the associated caption.

\clearpage

\subsection{Applications to multi-task learning}
\begin{sidewaysfigure}[!ht]
    \centering
    \includegraphics[width=\textwidth]{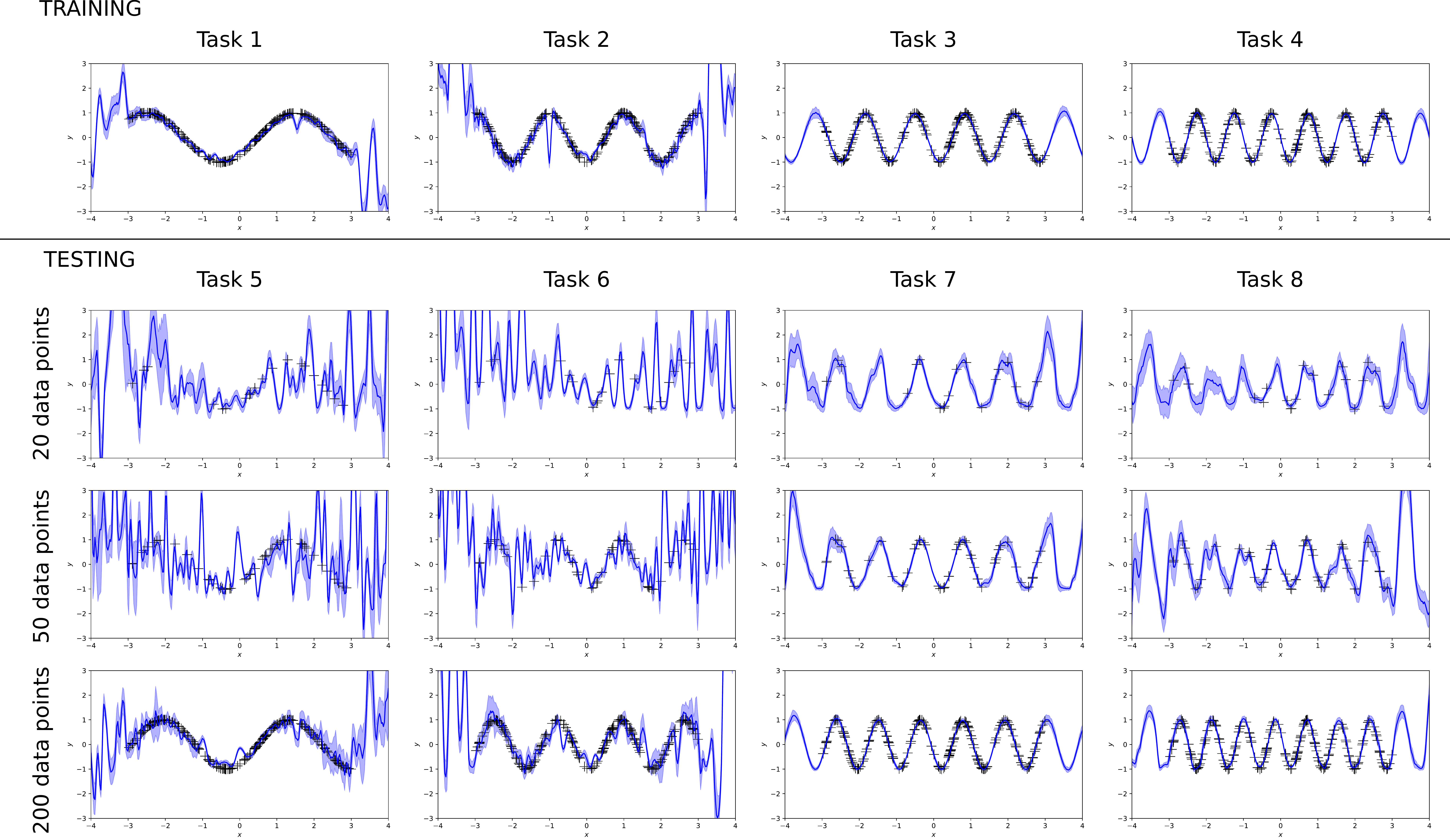}
    \caption{Training on multiple related tasks and adaptation to novel tasks at test time. In this case, the latent variables (as well as weight code hyperparameters) are shared across tasks while each individual has its own input-dependent kernel. At test time, only the private parameters for the new task are re-initialised and optimised. Best viewed in colour.}
    \label{fig:sin_multitask}
\end{sidewaysfigure}

We further investigate using the proposed model for multi-task learning. In particular, the latent variable $\zvec$ and corresponding hyper-parameters and variational parameters can be shared across different tasks whilst the meta mapping and the input-dependent kernel are private to each individual task. We first train the model on four regression tasks, each corresponds to a sinusoid of a particular frequency. At test time, a novel test set is shown to the model. The  hyper-parameters of the input kernel and variational parameters corresponding to this new test set are optimised while other hyper-parameters and the latent variables are kept fixed. We evaluate the performance of the model on the novel test sets to see how the latent variables can be reused and shared across tasks to facilitate fast adaptation to new settings. The performance of the model on the tasks used for training and new tasks at test time is shown in \cref{fig:sin_multitask}. This result demonstrates the ability of the model trained with multiple similarly related tasks to faithfully and quickly adapt to new settings. 

\end{document}